\documentclass[journal]{IEEEtran}

\usepackage{silence}
\usepackage{caption}
\usepackage{subcaption}
\usepackage{multirow}
\usepackage{array}
\usepackage{booktabs}
\usepackage{amsmath}
\usepackage{amssymb}
\usepackage[nolist,nohyperlinks]{acronym}
\usepackage[inline,shortlabels]{enumitem}
\usepackage{siunitx}
\usepackage{tikz}
\usepackage{pgfplots}

\usepackage{etoolbox}
\usepackage{soul}
\renewcommand{\hl}[1]{#1}

\usepackage[hidelinks]{hyperref}

\usepackage{orcidlink}

\DeclareSIUnit{\pixel}{pixel}
\DeclareSIUnit{\pixels}{pixels}
\DeclareSIUnit{\band}{bands}
\DeclareSIUnit{\sband}{spectral \ bands}
\DeclareSIUnit{\patch}{patch}
\DeclareSIUnit{\patches}{patches}
\DeclareSIUnit{\tiles}{tiles}
\DeclareSIUnit{\epoch}{epochs}
\DeclareSIUnit{\megapixel}{MP}
\DeclareSIUnit{\bit}{bit}
\DeclareSIUnit{\day}{days}
\DeclareSIUnit{\flops}{FLOPS}
\DeclareSIUnit{\nothing}{\relax}

\usetikzlibrary{arrows}
\usetikzlibrary{positioning}
\usetikzlibrary{calc}
\usetikzlibrary{shapes.geometric}
\usetikzlibrary{patterns}
\usetikzlibrary{patterns.meta}
\usetikzlibrary{spy}
\usetikzlibrary{fit}
\usetikzlibrary{trees}

\tikzset{
    align=center,
    node distance=5mm and 5mm,
    mark size=2.5pt,
}

\tikzstyle{every node} = [very thick]
\tikzstyle{arrow} = [thick,draw,->,>=stealth]
\tikzstyle{arrow-reverse} = [thick,draw,<-,>=stealth]
\tikzstyle{add} = [circle,draw,inner sep=1.5pt]
\tikzstyle{block} = [rectangle,rounded corners=1pt,draw,minimum width=16mm,minimum height=8mm]
\tikzstyle{conv1x1} = [trapezium,draw,shape border rotate=270,trapezium angle=78,fill=blue!20]
\tikzstyle{conv3x3} = [trapezium,draw,shape border rotate=270,trapezium angle=78,fill=red!20]
\tikzstyle{act} = [rectangle,rounded corners=1pt,draw,minimum width=20mm,minimum height=20mm,fill=gray!20]
\tikzstyle{dim} = [ellipse,draw,densely dashed,minimum width=2cm,minimum height=2mm]

\tikzstyle{square} = [rectangle,rounded corners=1pt,minimum width=1mm,minimum height=1mm]
\tikzstyle{token} = [rectangle,rounded corners=2mm,minimum width=8mm,minimum height=5mm,draw,anchor=west,fill=magenta]
\tikzstyle{postoken} = [rectangle,rounded corners=2mm,minimum width=6mm,minimum height=1mm,draw,anchor=west,fill=cyan]

\pgfplotsset{
    compat=newest,
    grid=both,
    grid style={line width=.1pt, draw=gray!10},
    major grid style={line width=.2pt,draw=gray!50},
    every axis plot/.append style={line width=0.8pt},
}

\newcommand{\matr}[1]{\mathbf{#1}} 
\newcommand{\vect}[1]{\mathbf{#1}}

\newcolumntype{H}{>{\setbox0=\hbox\bgroup}c<{\egroup}@{}}

\makeatletter
\def\@citex[#1]#2{\leavevmode
\let\@citea\@empty
\@cite{\@for\@citeb:=#2\do
{\@citea\def\@citea{,\penalty\@m}%
\edef\@citeb{\expandafter\@firstofone\@citeb\@empty}%
\if@filesw\immediate\write\@auxout{\string\citation{\@citeb}}\fi
\@ifundefined{b@\@citeb}{\hbox{\reset@font\bfseries ?}%
\G@refundefinedtrue
\@latex@warning
{Citation `\@citeb' on page \thepage \space undefined}}%
{\@cite@ofmt{\csname b@\@citeb\endcsname}}}}{#1}}
\makeatother

\robustify\ac
\robustify\acs
\robustify\acp
\robustify\acsp
\robustify\acf
\robustify\acfp
\robustify\qtyproduct
\robustify\num

\title{Adjustable Spatio-Spectral Hyperspectral Image Compression Network}

\author{%
    Martin Hermann Paul Fuchs\,\orcidlink{0009-0003-7800-284X},
    Behnood Rasti\,\orcidlink{0000-0002-1091-9841},~\IEEEmembership{Senior Member,~IEEE},
    and~Begüm Demir\,\orcidlink{0000-0003-2175-7072},~\IEEEmembership{Senior Member,~IEEE}%
    \thanks{Martin Hermann Paul Fuchs, Behnood Rasti and Begüm Demir are with the Faculty of Electrical Engineering and Computer Science, Technische Universität Berlin, 10623 Berlin, Germany and also with the \acf{bifold}, 10623 Berlin, Germany (e-mail: m.fuchs@tu-berlin.de; behnood.rasti@tu-berlin.de; demir@tu-berlin.de).}%
}

\begin{document}

\begin{acronym}
    \acro{ours}[HyCASS]{\textbf{A}djustable \textbf{S}patio-\textbf{S}pectral \textbf{Hy}perspectral Image \textbf{C}ompression Network}

    \acro{enmap}[EnMAP]{Environmental Mapping and Analysis Program}
    \acro{ohid1}[OHID-1]{Orbita Hyperspectral Images Dataset-1}
    \acro{l2a}[L2A]{Level 2A}

    \acro{rs}[RS]{remote sensing}
    \acro{eo}[EO]{earth obervation}
    \acro{cv}[CV]{computer vision}
    \acro{gpu}[GPU]{graphics processing unit}
    \acro{rgb}[RGB]{red, green and blue}
    \acro{hsi}[HSI]{hyperspectral image}
    \acro{cnn}[CNN]{convolutional neural network}
    \acro{ann}[ANN]{artificial neural network}
    \acro{spiht}[SPIHT]{set partitioning in hierarchical trees}
    \acro{speck}[SPECK]{set partitioning embedded block}
    \acro{pca}[PCA]{principle component analysis}
    \acro{dct}[DCT]{discrete cosine transform}
    \acro{klt}[KLT]{Karhunen–Loève transform}
    \acro{se}[SE]{Squeeze and Excitation}
    \acro{swin}[Swin]{Shifted windows}
    \acro{rstb}[RSTB]{Residual Swin Transformer Block}
    \acro{ntu}[NTU]{Neural Transformation Unit}
    \acro{fe}[FE]{feature embedding}
    \acro{fu}[FU]{feature unembedding}
    \acro{stl}[STL]{Swin Transformer layer}
    \acro{wa}[WA]{window attention}
    \acro{swa}[SWA]{shifted window attention}
    \acro{fm}[FM]{foundation model}
    
    \acro{psnr}[PSNR]{peak signal-to-noise ratio}
    \acro{sa}[SA]{spectral angle}
    \acro{mse}[MSE]{mean squared error}
    \acro{ssim}[SSIM]{structural similarity index measure}
    \acro{cr}[CR]{compression ratio}
    \acro{bpppc}[bpppc]{bits per pixel per channel}
    \acro{decibel}[\si{\decibel}]{decibels}
    \acro{gsd}[GSD]{ground sample distance}
    \acro{flops}[FLOPs]{floating point operations}
    \acro{lr}[LR]{learning rate}
    \acro{bs}[BS]{batch size}

    \acro{leakyrelu}[LeakyReLU]{leaky rectified linear unit}
    \acro{prelu}[PReLU]{parametric rectified linear unit}

    \acro{dpcm}[DPCM]{Differential Pulse Code Modulation}
    \acro{ae}[AE]{autoencoder}
    \acro{vae}[VAE]{variational autoencoder}
    \acro{cae}[CAE]{convolutional autoencoder}
    \acro{gan}[GAN]{generative adversarial network}
    \acro{inr}[INR]{Implicit Neural Representations}
    
    \acro{a1dcae}[A1D-CAE]{Adaptive 1-D Convolutional Autoencoder}
    \acro{1dcae}[1D-CAE]{1D-Convolutional Autoencoder}
    \acro{sscnet}[SSCNet]{Spectral Signals Compressor Network}
    \acro{3dcae}[3D-CAE]{3D Convolutional Auto-Encoder}
    \acro{linerwkv}[LineRWKV]{Line Receptance Weighted Key Value}
    \acro{hycot}[HyCoT]{Hyperspectral Compression Transformer}
    \acro{tic}[TIC]{Transformer-based Image Compression}
    \acro{s2cnet}[S2C-Net]{Spatio-Spectral Compression Network}
    \acro{hific}[HiFiC]{High Fidelity Compression}

    \acro{mlp}[MLP]{multilayer perceptron}

    \acro{1d}[1D]{one-dimensional}
    \acro{2d}[2D]{two-dimensional}
    \acro{3d}[3D]{three-dimensional}

    \acro{ln}[LN]{layer normalization}

    \acro{rsim}[RSiM]{Remote Sensing Image Analysis}
    \acro{bifold}[BIFOLD]{Berlin Institute for the Foundations of Learning and Data}
\end{acronym}

\maketitle

\begin{abstract}
With the rapid growth of hyperspectral data archives in \ac{rs}, the need for efficient storage has become essential, driving significant attention toward learning-based \ac{hsi} compression.
However, a comprehensive investigation of the individual and joint effects of spectral and spatial compression on learning-based \ac{hsi} compression has not been thoroughly examined yet.
Conducting such an analysis is crucial for understanding how the exploitation of spectral, spatial, and joint spatio-spectral redundancies affects \ac{hsi} compression.
To address this issue, we propose \ac{ours}, a learning-based model designed for adjustable \ac{hsi} compression in both spectral and spatial dimensions.
\ac{ours} consists of six main modules:
\begin{enumerate*}[1)]
    \item spectral encoder \hl{module};
    \item spatial encoder \hl{module};
    \item \ac{cr} adapter encoder \hl{module};
    \item \ac{cr} adapter decoder \hl{module};
    \item spatial decoder \hl{module}; and
    \item spectral decoder \hl{module}.
\end{enumerate*}
The modules employ convolutional layers and transformer blocks to capture both short-range and long-range redundancies.
Experimental results on \hl{three} \ac{hsi} benchmark datasets demonstrate the effectiveness of our proposed adjustable model compared to existing learning-based compression models\hl{, surpassing the state of the art by up to  {\SI[round-mode=places,round-precision=2,detect-weight=true]{2.364}{\decibel}} in terms of {\acs{psnr}}}.
Based on our results, we establish a guideline for effectively balancing spectral and spatial compression across different \acp{cr}, taking into account the spatial resolution of the \acp{hsi}.
Our code and pre-trained model weights are publicly available at \url{https://git.tu-berlin.de/rsim/hycass}.
\end{abstract}

\acresetall

\begin{IEEEkeywords}
Hyperspectral image compression, adjustable spatio-spectral compression, deep learning, remote sensing.
\end{IEEEkeywords}

\IEEEpeerreviewmaketitle

\section{Introduction}
\label{sec:introduction}
Hyperspectral sensors capture images spanning hundreds of continuous bands across the electromagnetic spectrum.
Fine spectral information provided in \acp{hsi} enables the identification and differentiation of materials within a scene.
Hyperspectral sensors, mounted on satellites, airplanes, and drones, enable a wide range of \ac{rs} applications \hl{{\cite{cheng2025modern}}}, including forest monitoring \cite{lin2024model}, water quality assessment \cite{fabbretto2024tracking}, wildfire detection \cite{spiller2023wildfire}, flood mapping \cite{masari2024manifold}\hl{, yield estimation {\cite{liu2025utilizing}} and plastic litter detection {\cite{balsi2025plastic}}}.
The continuous improvement in hyperspectral sensors has enabled them to extract increasingly detailed spectral and spatial information, which is essential for advanced analysis. However, the substantial volume of data produced by these sensors presents significant challenges in terms of storage, transmission, and processing.
An emerging research area focuses on the efficient compression of hyperspectral data to preserve crucial spectral and spatial information content for subsequent analysis \cite{gomes2025lossy}.

Many \ac{hsi} compression methods are presented in the literature.
Generally, they can be categorized into three classes:
\begin{enumerate*}[i)]
    \item lossless;
    \item near-lossless; and
    \item lossy
\end{enumerate*}
\ac{hsi} compression.
Each category offers a unique trade-off between data preservation and compression efficiency.
Lossless \ac{hsi} compression ensures a perfect reconstruction of the original data without any loss of information and is therefore particularly important for tasks with zero tolerance for data degradation.
However, lossless \ac{hsi} compression methods typically only achieve \acp{cr} of \SIrange{2}{4}{} \cite{altamimi2024lossless}.
This restriction limits their applicability in scenarios with strict bandwidth or storage space constraints that require significantly higher \acp{cr}.
Near-lossless \ac{hsi} compression achieves higher \acp{cr} than lossless compression while introducing minimal distortion between the original and reconstructed \acp{hsi}.
The maximum error is upper-bounded, ensuring controlled deviation in the reconstructed \acp{hsi}.
However, small errors may accumulate, potentially impacting applications that require absolute spectral precision.
Furthermore, the achievable \acp{cr} remain limited compared to lossy compression.
Lossy \ac{hsi} compression offers significant advantages, particularly in achieving high \acp{cr}, which is essential for applications with strict bandwidth or storage limitations \cite{gomes2025lossy}.
By selectively discarding less important information, lossy compression efficiently reduces data size while preserving key features, making it ideal for large-scale hyperspectral data transmission, storage, and processing.
Although lossy compression introduces some degree of information loss, the degradation is typically negligible and does not substantially compromise the usability of the data in most practical applications.

From a methodological point of view, \ac{hsi} compression can be grouped into two categories:
\begin{enumerate*}[i)]
    \item traditional methods; and
    \item learning-based methods.
\end{enumerate*}
Traditional \ac{hsi} compression methods have been extensively investigated in \ac{rs} with predictive coding emerging as a common approach \cite{dua2020comprehensive}.
Predictive coding takes advantage of both spectral and spatial redundancies by predicting pixel values based on contextual information and encoding only the residuals between predicted and actual values for efficient storage or transmission.
For example, in \cite{mielikainen2004clustered}, a clustered \ac{dpcm} compression method, which clusters spectra and calculates an optimized predictor for each cluster, is proposed for \acp{hsi}. After linear prediction of a spectrum, the difference is entropy-coded using an adaptive entropy coder for each of the clusters. 
Another prominent implementation of predictive coding is the CCSDS 123.0-B-2 standard \cite{hernandez2021ccsds} for lossless \ac{hsi} compression that employs adaptive linear prediction to minimize redundancies in \acp{hsi}. Its low complexity and flexible architecture, coupled with the capability to achieve near-lossless compression through closed-loop quantization, make it well-suited for deployment in onboard \ac{rs} systems.
Although prediction-based methods are widely used for lossless and near-lossless \ac{hsi} compression, they cannot generate meaningful latent representations, and their autoregressive functionality results in a slow processing speed.

In contrast, traditional transform coding methods excel at extracting compact latent representations from hyperspectral data and are frequently used for lossy compression.
They project hyperspectral data into a lower-dimensional, decorrelated latent space using mathematical transformations.
The resulting reduced number of coefficients is subsequently quantized, introducing some loss of information, and then entropy-coded.
Several traditional transform-coding methods are proposed in \ac{rs}.
As an example, in \cite{du2007hyperspectral}, \ac{pca} is combined with JPEG2000 for joint spatio-spectral compression of \acp{hsi}, whereas \ac{pca} is applied followed by \ac{dct} in \cite{yadav2018compression}. 
In \cite{dragotti2000compression}, \ac{3d} transform coding is achieved by applying wavelet transformation in the spatial dimensions and \ac{klt} in the spectral dimension, followed by \ac{3d}-\acs{spiht} for efficient lossy compression.
In \cite{tang2006three} \ac{3d}-\acs{speck}, which takes advantage of the \ac{3d} wavelet transform to efficiently encode \acp{hsi} by compressing the redundancies in all three dimensions, is introduced.
Despite their effectiveness in extracting compact features and achieving high \acp{cr}, traditional methods often rely on hand-crafted transformations, which limit their ability to fully exploit the rich spatio–spectral structure of hyperspectral data.
Consequently, these methods may not generalize well when applied to hyperspectral data captured under different conditions, sensor types, or scene characteristics.

To overcome these limitations, the development of learning-based \ac{hsi} compression has recently attracted great attention in \ac{rs}. Learning-based \ac{hsi} compression methods leverage deep neural networks to automatically learn hierarchical and data-adaptive representations from large-scale training datasets, allowing more effective characterization of complex spectral and spatial redundancies. These methods also lead to an improved rate–distortion performance and better generalization across different scenes \cite{gomes2025lossy}. As an example, in \cite{valsesia2024onboard} the \acs{linerwkv} method that enhances CCSDS 123.0-B-2 \cite{hernandez2021ccsds} by introducing a learning-based predictor is presented. \acs{linerwkv} achieves superior lossless and near-lossless reconstruction performance compared to CCSDS 123.0-B-2 at the cost of increased training time and computational complexity.
In \cite{fuchs2024generative}, the authors introduce two \ac{gan}-based models designed for spatio-spectral compression of \acp{hsi}. Their model extends the \ac{hific} framework \cite{mentzer2020high} by incorporating:
\begin{enumerate*}[i)]
    \item \ac{se} blocks; and
    \item \ac{3d} convolutions
\end{enumerate*}
to better exploit spectral redundancies alongside spatial redundancies. Although these models are capable of achieving extremely high \acp{cr}, exceeding $10^4$, their generative nature can lead to the synthesis of unrealistic spectral and spatial content, potentially compromising the fidelity of the reconstructed data.
\hl{In {\cite{mijares2023scalable}}, a channel clusterization strategy is proposed for onboard {\ac{hsi}} compression to reduce the computational demands and make the method scalable for different data sources with a varying number of spectral bands. While the method achieves high quality reconstructions for low {\acp{cr}}, its robustness is highly dependent on a proper tuning of the number of features and the latent space dimensionality.}
In \cite{rezasoltani2024hyperspectral}, a method for \ac{hsi} compression using \ac{inr} is presented, where an \ac{mlp} network learns to map pixel locations to pixel intensities. The weights of the learned model are stored or transmitted to achieve compression. 
In \cite{zhao2025paradigm}, a neural video representation approach is proposed. \acp{hsi} are treated as a stream of video data, where each spectral band represents a frame of information, and variances between spectral bands represent transformations between frames. Their approach utilizes the spectral band index and the spatial coordinate index as its input to perform network overfitting.
A fundamental limitation of neural representation approaches lies in their training procedure, which involves overfitting a separate neural network to each \ac{hsi}. This instance-specific optimization results in substantial computational costs, limiting the practicality of large-scale data processing.

Most state-of-the-art learning-based \ac{hsi} compression methods adopt the \ac{ae} architecture, where an encoder network compresses the input data into a compact latent representation, and a decoder network reconstructs the data from that. This structure enables end-to-end optimization and allows the networks to learn efficient, data-driven representations of spectral and spatial information. Existing \ac{ae} models differ primarily in how they process the spectral and spatial dimensions by using \ac{1d}, \ac{2d}, and \ac{3d} convolutional layers, to balance compression efficiency, reconstruction quality, and computational complexity.
As an example, in \cite{kuester20211d, kuester2023adaptive}, \ac{1dcae} is presented, which compresses the spectral content without considering spatial redundancies by stacking multiple blocks of \ac{1d} convolutions, pooling layers, and \ac{leakyrelu} activations. Although high-quality reconstructions can be achieved with this model, the pooling layers limit the achievable \acp{cr} to $2^n$, where $n$ is the number of poolings. Another limitation of this model is the increasing computational complexity with higher \acp{cr} due to the deeper network architecture.
In \cite{la2022hyperspectral}, \ac{sscnet} extracts spatial features via \ac{2d} convolutional kernels. \ac{2d} max pooling layers introduce spatial compression, while the final \ac{cr} is adapted via the latent channels within the bottleneck. Although \ac{sscnet} enables significantly higher \acp{cr}, this comes at the cost of spatially blurred image reconstructions.
In \cite{chong2021end}, \ac{3dcae} is introduced for joint compression of spatio-spectral redundancies via \ac{3d} convolutional kernels. Additionally, residual blocks allow gradients to flow through the network directly and improve the model's performance. However, the \ac{3d} kernels significantly increase computational complexity.
In \cite{sprengel2024learning}, \ac{s2cnet} is introduced. Initially, a pixelwise \ac{ae} is pre-trained to capture the essential spectral features of the hyperspectral data. To enhance spatial redundancy removal, a spatial \ac{ae} network is added to the bottleneck layer of the spectral \ac{ae}. This dual-layer architecture allows the model to learn both spectral and spatial representations effectively. The entire model is then trained using a mixed loss function that combines reconstruction errors from both the spectral and spatial components. Although this model achieves state-of-the-art performance for high \acp{cr}, it falls short in optimally balancing the trade-off between spectral and spatial \ac{cr}. This limitation suggests room for improvement in achieving more balanced compression across both dimensions.
In \cite{fuchs2024hycot}, the authors propose \ac{hycot}, a transformer-based \ac{ae} designed for pixelwise compression of \acp{hsi} that exploits long-range spectral redundancies. The model significantly reduces computational complexity through random sampling and independent pixelwise processing, without notable degradation in reconstruction quality. However, similar to other spectral compression models, \ac{hycot} does not exploit spatial redundancies, thus limiting the achievable \acp{cr}.

Existing learning-based \ac{hsi} compression models have limitations to effectively address several fundamental challenges associated with hyperspectral data, including varying spatial resolutions, high spectral dimensionality, and the need for adjustability across a broad range of \acp{cr}. In particular, these models often lack mechanisms to dynamically exploit spectral and spatial redundancies under specific compression requirements and sensor characteristics. As a result, current approaches face limitations in terms of scalability, generalization, and their ability to balance compression efficiency, reconstruction fidelity, and computational complexity.

To overcome the above-mentioned critical issues, in this paper, we introduce \ac{ours}.
The proposed model aims to enable flexible spatio-spectral compression.
To this end, \ac{ours} employs six modules:
\begin{enumerate*}[i)]
    \item a spectral encoder module;
    \item a spatial encoder module;
    \item a \ac{cr} adapter encoder module;
    \item a \ac{cr} adapter decoder module;
    \item a spatial decoder module; and
    \item a spectral decoder module.
\end{enumerate*}
The novelty of the proposed model consists of the following:
\begin{enumerate*}[1)]
    \item a spectral feature extraction that captures spectral redundancies across the whole spectrum of each pixel independently (realized within the spectral encoder module and spectral decoder module);
    \item an adjustable number of spatial stages exploiting both short- and long-range spatial redundancies to control spatial compression (realized within the spatial encoder module and spatial decoder module); and
    \item an adjustable latent channel dimension to regulate spectral compression (realized within the \ac{cr} adapter encoder module and \ac{cr} adapter decoder module).
\end{enumerate*}
Unlike existing methods in the literature, our proposed model supports flexible compression in both spectral and spatial dimensions.
We evaluate the proposed model on \hl{three} distinct datasets (HySpecNet-11k \cite{fuchs2023hyspecnet}\hl{, Berlin-Urban-Gradient {\cite{okujeni2016berlin}}} and MLRetSet \cite{omruuzun2024novel}), demonstrating its effectiveness in compressing over a broad range of \acp{cr} and \hl{multiple} spatial resolutions.
Our experimental results show that spectral compression is preferable in cases where the \ac{cr} or spatial resolution is low, whereas spatio-spectral compression is more effective for \hl{medium and high} \acp{cr} or spatial resolutions.
The main contributions of this paper are summarized as follows:
\begin{itemize}
    \item We propose a spatio-spectral \ac{hsi} compression model with adjustable spatial stages and latent channels, capable of effective \ac{hsi} compression across a broad range of \acp{cr} and varying spatial resolutions.
    \item We provide extensive experimental results on \hl{three} benchmark datasets for the introduced model, including an ablation study, comparisons with other approaches, and visual analyses of the reconstructions.
    \item We analyze the effects of spectral and spatial \ac{hsi} compression on the reconstruction quality for multiple \acp{cr} and \hl{three} \acp{gsd}, providing a comprehensive evaluation of their individual and combined impacts on reconstruction performance.
    \item We demonstrate the advantages of adjustable deep learning architectures and derive guidelines for the trade-off between spectral and spatial compression under varying \ac{cr} conditions, for \hl{low, medium, and high} spatial resolution hyperspectral data.
\end{itemize}
The remainder of this paper is organized as follows:
\autoref{sec:proposed-method} introduces the proposed \ac{ours} model.
\autoref{sec:dataset-description-and-experimental-setup} describes the considered datasets and provides the design of experiments, while the experimental results are presented in \autoref{sec:experimental-results}.
Finally, in \autoref{sec:conclusion}, the conclusion of the work is drawn.

\section{Proposed Adjustable Spatio-Spectral Hyperspectral Image
Compression Network (H\textnormal{y}CASS)}
\label{sec:proposed-method}

Let $\matr{X} \in \mathbb{R}^{H \times W \times C}$ denote an \ac{hsi} with spatial dimensions $H$ and $W$, and $C$ spectral bands.
In this work, we focus on lossy \ac{hsi} compression that transforms the original \ac{hsi} $\matr{X}$ into a compact and decorrelated latent representation $\matr{Y} \in \mathbb{R}^{\Sigma \times \Omega \times \Gamma}$.
Here, $\Sigma$ and $\Omega$ denote the reduced latent spatial dimensions, while $\Gamma$ represents the number of latent channels.
The latent representation $\matr{Y}$ should retain sufficient information, such that the original \ac{hsi} can be approximately reconstructed from $\matr{Y}$ as $\matr{\hat{X}} \in \mathbb{R}^{H \times W \times C}$.
The compression aims to minimize the distortion $d: \matr{X} \times \matr{\hat{X}} \rightarrow [0, \infty)$ between $\matr{X}$ and $\matr{\hat{X}}$ for a fixed \ac{cr}.

\begin{figure*}
    \centering
    \begin{tikzpicture}
        \node[inner sep=0pt] (enc-in) {\includegraphics[width=.118\linewidth]{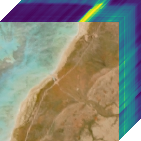}};

        \node[conv1x1,right=of enc-in] (enc-conv-0) {Conv2D \\ $C \rightarrow N$ \\ \qtyproduct{1x1}{}};
        \node[act,right=of enc-conv-0,minimum width=5mm] (enc-act-0) {\rotatebox{90}{\acs{leakyrelu}}};

        \node[right=of enc-act-0,inner sep=0pt,outer sep=0pt] (enc-tmp-0) {$\bullet$};
        
        \node[block,right=of enc-tmp-0,fill=teal!20] (enc-swin) {Residual \\ \acs{swin} \\ Transformer \\ Block};
        \node[conv3x3,right=of enc-swin] (enc-conv2d) {Conv2D $\downarrow^2$ \\ $N \rightarrow N$ \\ \qtyproduct{3x3}{}};
        \node[act,right=of enc-conv2d,minimum width=5mm] (enc-act) {\rotatebox{90}{\acs{leakyrelu}}};

        \node[right=of enc-act,inner sep=0pt,outer sep=0pt] (enc-tmp-1) {$\bullet$};
        
        \node[conv1x1,right=of enc-tmp-1] (enc-conv-1) {Conv2D \\ $N \rightarrow \Gamma$ \\ \qtyproduct{1x1}{}};
        \node[act,right=of enc-conv-1,minimum width=5mm] (enc-act-1) {\rotatebox{90}{Sigmoid}};
        
        \node[dim,below right=of enc-act-1,xshift=-12mm] (lat) {Latent\\Representation};

        \node[conv1x1,below left=of lat,xshift=12mm,yshift=-3.2mm] (dec-conv-1) {Conv2D \\ $\Gamma \rightarrow N$ \\ \qtyproduct{1x1}{}};
        \node[act,left=of dec-conv-1,minimum width=5mm] (dec-act-1) {\rotatebox{90}{\acs{leakyrelu}}};

        \node[left=of dec-act-1,inner sep=0pt,outer sep=0pt] (dec-tmp-1) {$\bullet$};
        
        \node[conv3x3,left=of dec-tmp-1] (dec-conv2d) {Conv2D $\uparrow^2$ \\ $N \rightarrow N$ \\ \qtyproduct{3x3}{}};
        \node[act,left=of dec-conv2d,minimum width=5mm] (dec-act) {\rotatebox{90}{\acs{leakyrelu}}};
        \node[block,left=of dec-act,fill=teal!20] (dec-swin) {Residual \\ \acs{swin} \\ Transformer \\ Block};

        \node[left=of dec-swin,inner sep=0pt,outer sep=0pt] (dec-tmp-0) {$\bullet$};
        
        \node[conv1x1,left=of dec-tmp-0] (dec-conv-0) {Conv2D \\ $N \rightarrow C$ \\ \qtyproduct{1x1}{}};
        \node[act,left=of dec-conv-0,minimum width=5mm] (dec-act-0) {\rotatebox{90}{Sigmoid}};
        
        \node[inner sep=0pt,left=of dec-act-0] (dec-out) {\includegraphics[width=.118\linewidth]{img/hsi.pdf}};

        \draw[arrow] (enc-in) -- (enc-conv-0);
        \draw[arrow] (enc-conv-0) -- (enc-act-0);
        \draw[arrow] (enc-act-0) -- (enc-swin);
        \draw[arrow] (enc-swin) -- (enc-conv2d);
        \draw[arrow] (enc-conv2d) -- (enc-act);
        \draw[arrow] (enc-act) -- (enc-conv-1);
        \draw[arrow] (enc-conv-1) -- (enc-act-1);

        \draw[arrow] (enc-act-1) -| (lat);
        \draw[arrow] (lat) |- (dec-conv-1);

        \draw[arrow] (dec-conv-1) -- (dec-act-1);
        \draw[arrow] (dec-act-1) -- (dec-conv2d);
        \draw[arrow] (dec-conv2d) -- (dec-act);
        \draw[arrow] (dec-act) -- (dec-swin);
        \draw[arrow] (dec-swin) -- (dec-conv-0);
        \draw[arrow] (dec-conv-0) -- (dec-act-0);
        \draw[arrow] (dec-act-0) -- (dec-out);

        \draw[arrow] (enc-tmp-1) -- ++(0,1.5) -| node[pos=0.25, above] (enc-N) {$S \times$} (enc-tmp-0);
        \draw[arrow] (dec-tmp-0) -- ++(0,-1.5) -| node[pos=0.25, below] (dec-N) {$S \times$} (dec-tmp-1);

        \node[draw=cyan,dotted,fit=(enc-conv-0)(enc-act-0),inner sep=2mm,label=above:{\textcolor{cyan}{\shortstack{Spectral \\ Encoder Module}}}] {};

        \node[draw=cyan,dotted,fit=(enc-conv-1)(enc-act-1),inner sep=2mm,label=above:{\textcolor{cyan}{\shortstack{\acs{cr} Adapter \\ Encoder Module}}}] {};

        \node[draw=cyan,dotted,fit=(enc-swin)(enc-conv2d)(enc-act)(enc-tmp-0)(enc-tmp-1)(enc-N),inner sep=2mm,label=above:{\textcolor{cyan}{Spatial Encoder Module}}] {};

        \node[draw=magenta,dotted,fit=(dec-conv-0)(dec-act-0),inner sep=2mm,label=below:{\textcolor{magenta}{\shortstack{Spectral \\ Decoder Module}}}] {};

        \node[draw=magenta,dotted,fit=(dec-conv-1)(dec-act-1),inner sep=2mm,label=below:{\textcolor{magenta}{\shortstack{\acs{cr} Adapter \\ Decoder Module}}}] {};

        \node[draw=magenta,dotted,fit=(dec-swin)(dec-conv2d)(dec-act)(dec-tmp-0)(dec-tmp-1)(dec-N),inner sep=2mm,label=below:{\textcolor{magenta}{Spatial Decoder Module}}] {};
        
    \end{tikzpicture}
    \caption{Overview of our proposed \ac{ours} model. Initially, a pixelwise convolution in the spectral encoder module extracts spectral features. The spatial encoder module, composed of $S \times$ stacked stages, performs both long- and short-range spatial feature extraction, where each spatial stage introduces higher spatial compression. Subsequently, the \ac{cr} adapter encoder module adjusts the size of the latent representation to match the targeted spatio-spectral \ac{cr}. The decoder mirrors the encoder structure, replacing downsampling with upsampling operations.}
    \label{fig:block-diagram}
\end{figure*}

To enable effective spatio-spectral \ac{hsi} compression, we propose \ac{ours}.
The proposed model combines pixelwise convolutions, strided \ac{2d} convolutional layers, and Residual \ac{swin} Transformer Blocks \cite{liu2021swin} to leverage both short-range and long-range redundancies across the spectral as well as the spatial dimension of \acp{hsi}.
As illustrated in \autoref{fig:block-diagram}, \ac{ours} consists of six modules within both the encoder and decoder:
\begin{enumerate*}[i)]
    \item a spectral encoder module that involves a spectral feature extraction;
    \item a spatial encoder module with a configurable number of spatial stages $S$ for adjustable spatial compression, incorporating short-range and long-range spatial redundancies;
    \item a \ac{cr} adapter encoder module to balance the trade-off between compressing spectral and spatial information content via the number of latent channels $\Gamma$, depending on the joint spatio-spectral target \ac{cr};
    \item a \ac{cr} adapter decoder module that recovers spectral information;
    \item a spatial decoder module that performs spatial reconstruction; and
    \item a spectral decoder module for spectral reconstruction.
\end{enumerate*}

\hl{In summary, the functional flow of {\ac{ours}} is as follows:
First, the spectral encoder module extracts spectral redundancies, which are subsequently processed by the spatial encoder module to incorporate spatial redundancies.
This joint representation is then further spectrally compressed by the {\ac{cr}} adapter encoder module, forming a low-dimensional latent representation.
On the decoder side, the latent representation is then spectrally expanded through the CR adapter decoder module, followed by a reconstruction of spatial information via the spatial decoder module. Finally, the spectral decoder module reconstructs spectral information.}

\ac{ours} facilitates adjustable spatio-spectral compression, striking a balance between reconstruction fidelity and compression efficiency.
This balance is achieved through the adjustable parameters $\Gamma$ and $S$, which control spectral and spatial \ac{cr}, respectively.
\hl{In the following subsections, we provide detailed explanations of the six modules comprising {\ac{ours}}.}

\subsection{HyCASS Spectral Encoder Module}
Initially, the spectral encoder module $E_\Phi: \mathbb{R}^{H \times W \times C} \rightarrow \mathbb{R}^{H \times W \times N}$ of the proposed \ac{ours} model, which is defined as:
\begin{align}
    E_\Phi \left( \matr{\xi} \right) = \text{\acs{leakyrelu}} \left( \text{Conv2D}_{\qtyproduct{1x1}{}}^{C \rightarrow N} \left( \matr{\xi} \right) \right)
\end{align}
performs spectral feature extraction using a pixelwise convolution, realized via a \qtyproduct{1x1}{} kernel inside a \ac{2d} convolutional layer.
Although \ac{2d} convolutions are typically used for capturing spatial patterns, the use of a \qtyproduct{1x1}{} kernel ensures that the convolution is applied independently to each pixel location without aggregating any spatial context.
The pixelwise convolution captures spectral redundancies along the whole spectrum of each pixel by projecting the high-dimensional number $C$ of spectral bands into a decorrelated, spectral representation with $N$ channels.
A \ac{leakyrelu} is applied after the convolutional layer to introduce non-linear activation, thereby enhancing its capacity to learn complex patterns from data.
This enables the spectral features extracted from the previous layer to capture more complex relationships in the hyperspectral data.
The extracted spectral features are subsequently processed by the spatial encoder module.

\subsection{HyCASS Spatial Encoder Module}
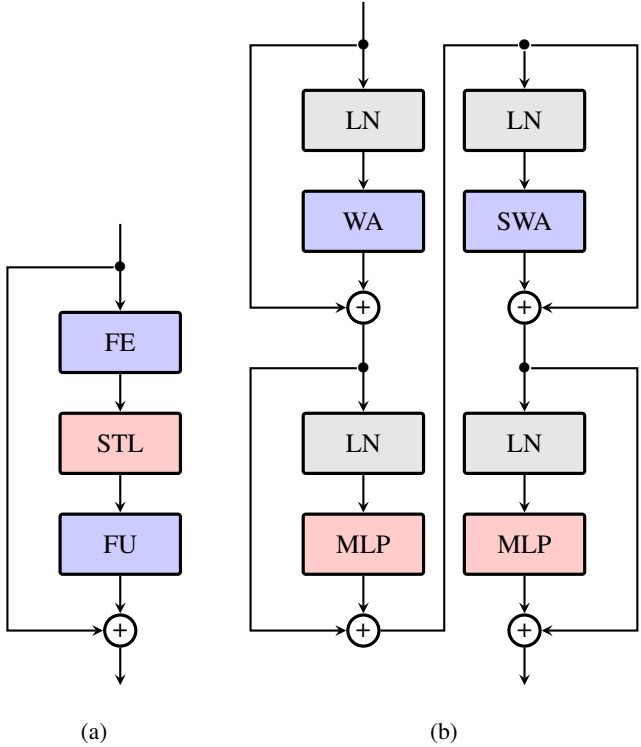
\begin{figure}
    \centering
    \begin{subfigure}[t]{0.31\linewidth}
        \centering
        \begin{tikzpicture}
            \node (in) {};
            \node[below=of in,inner sep=0pt,outer sep=0pt] (res) {$\bullet$};
            \node[block,below=of res,fill=blue!20] (fe) {\acs{fe}};
            \node[block,below=of fe,fill=red!20] (stl) {\acs{stl}};
            \node[block,below=of stl,fill=blue!20] (fu) {\acs{fu}};
            \node[add,below=of fu] (add) {+};
            \node[below=of add] (out) {};

            \draw[arrow] (in) -- (fe);
            \draw[arrow] (fe) -- (stl);
            \draw[arrow] (stl) -- (fu);
            \draw[arrow] (fu) -- (add);
            \draw[arrow] (add) -- (out);

            \draw[arrow] (res) -- ++(-1.5,0) |- (add);
        \end{tikzpicture}
        \caption{}
        \label{fig:rstb-rstb}
    \end{subfigure}
    \hfill
    \begin{subfigure}[t]{0.64\linewidth}
        \centering
        \begin{tikzpicture}
            \node (in) {};
            \node[below=of in,inner sep=0pt,outer sep=0pt] (res-1) {$\bullet$};
            \node[block,below=of res-1,fill=gray!20] (ln-1) {\acs{ln}};
            \node[block,below=of ln-1,fill=blue!20] (wa) {\acs{wa}};
            \node[add,below=of wa] (add-1) {+};

            \node[below=of add-1,inner sep=0pt,outer sep=0pt] (res-2) {$\bullet$};
            \node[block,below=of res-2,fill=gray!20] (ln-2) {\acs{ln}};
            \node[block,below=of ln-2,fill=red!20] (mlp-1) {\acs{mlp}};
            \node[add,below=of mlp-1] (add-2) {+};

            \node[block,right=of ln-1,fill=gray!20] (ln-3) {\acs{ln}};
            \node[above=of ln-3,inner sep=0pt,outer sep=0pt] (res-3) {$\bullet$};
            \node[block,below=of ln-3,fill=blue!20] (swa) {\acs{swa}};
            \node[add,below=of swa] (add-3) {+};

            \node[below=of add-3,inner sep=0pt,outer sep=0pt] (res-4) {$\bullet$};
            \node[block,below=of res-4,fill=gray!20] (ln-4) {\acs{ln}};
            \node[block,below=of ln-4,fill=red!20] (mlp-2) {\acs{mlp}};
            \node[add,below=of mlp-2] (add-4) {+};

            \node[below=of add-4] (out) {};

            \draw[arrow] (in) -- (ln-1);
            \draw[arrow] (ln-1) -- (wa);
            \draw[arrow] (wa) -- (add-1);
            \draw[arrow] (add-1) -- (ln-2);
            \draw[arrow] (ln-2) -- (mlp-1);
            \draw[arrow] (mlp-1) -- (add-2);

            \draw[arrow] (ln-3) -- (swa);
            \draw[arrow] (swa) -- (add-3);
            \draw[arrow] (add-3) -- (ln-4);
            \draw[arrow] (ln-4) -- (mlp-2);
            \draw[arrow] (mlp-2) -- (add-4);
            \draw[arrow] (add-4) -- (out);

            \draw[arrow] (res-1) -- ++(-1.5,0) |- (add-1);
            \draw[arrow] (res-2) -- ++(-1.5,0) |- (add-2);

            \draw[arrow] (res-3) -- ++(1.5,0) |- (add-3);
            \draw[arrow] (res-4) -- ++(1.5,0) |- (add-4);

            \draw[arrow] (add-2) -- ++(1.075,0) |- (res-3) -- (ln-3);
            
        \end{tikzpicture}
        \caption{}
        \label{fig:rstb-stl}
    \end{subfigure}
    \caption{Architecture of (\subref{fig:rstb-rstb}) \acs{rstb} and (\subref{fig:rstb-stl}) \acs{stl}. \hl{The {\acs{rstb}} captures long-range spatial redundancies using {\acs{fe}}, {\acs{stl}} and {\acs{fu}} subcomponents. The {\acs{stl}} applies multi-head self-attention within and aross local windows using {\acs{ln}}, {\acs{wa}}, {\acs{swa}} and {\acs{mlp}} subcomponents.} Layout is redesigned based on \cite{liu2021swin} and   \cite{lu2021transformer}.}
    \label{fig:rstb}
\end{figure}

Following the spectral encoding, \ac{ours} applies a spatial encoder module to capture and compress spatial redundancies within the hyperspectral data.
The \ac{ours} spatial encoder $E_\chi: \mathbb{R}^{H \times W \times N} \rightarrow \mathbb{R}^{\Sigma \times \Omega \times N}$ defined as:
\begin{align}
    \begin{cases}
        E_\chi \left( \matr{\xi} \right) = f^{\left( S \right)} \left( \matr{\xi} \right), \quad S \in \mathbb{N}_0 \\
        f  \left( \matr{\xi} \right) = \text{\acs{leakyrelu}} \left( \text{Conv2D}_{\qtyproduct{3x3}{}}^{\downarrow^2} \left( \text{\acs{rstb}} \left( \matr{\xi} \right) \right) \right)
    \end{cases}
\end{align}
consists of a configurable sequence of spatial stages, denoted as $S$, each implemented as a \ac{ntu} \cite{lu2021transformer}.
This systematically reduces the spatial dimension while enriching the feature representation with contextual spatial information.
$S$ is a configurable hyperparameter that determines the spatial \ac{cr}, given by $\ac{cr}_\text{spat} = 4^S$.
This hyperparameter helps to adjust spatial compression to align with the targeted spatio-spectral \ac{cr}.
Each \ac{ntu} performs stepwise spatial compression and is composed of three main components:
\begin{enumerate*}[i)]
    \item a \ac{rstb} \cite{liu2021swin},
    \item a strided 2D convolutional layer with a kernel size of \qtyproduct{3x3}{}, and
    \item a \ac{leakyrelu} activation function.
\end{enumerate*}
The \ac{rstb} employs shifted window self-attention to capture long-range spatial redundancies as illustrated in \autoref{fig:rstb}.
It consists of three subcomponents:
\begin{enumerate*}[i)]
    \item \acf{fe} that reorders the input feature channels into a token sequence;
    \item \acf{stl} that applies multi-head self-attention within and across local windows using \ac{ln}, \ac{wa}, \ac{swa} and \ac{mlp}; and
    \item \acf{fu} that reorders the tokens back to their original spatial shape.
\end{enumerate*}
Residual connections mitigate the vanishing gradient issue, enhancing training stability.
This design ensures efficient attention computation while preserving spatial locality and contextual information.
Notably, patch division and linear embedding for tokenization are omitted as described in \cite{lu2021transformer}.
Following the \ac{rstb}, the strided \ac{2d} convolution captures local short-range spatial redundancies.
This layer downsamples the feature maps by a factor of \num{2} along both height and width, effectively reducing the image size by a factor of \num{4} per stage. The number of channels $N$ remains constant across all stages.
A non-linear \ac{leakyrelu} activation function is applied at the end of each \ac{ntu} to enhance the model's capacity to learn complex patterns.
This hierarchical spatial encoding progressively compresses the spatial content while preserving important structural and contextual details.
We would like to note that in the case of zero spatial stages ($S = \SI{0}{\times}$), \ac{ours} operates as spectral compression model while spatio-spectral compression is achieved with one or more spatial stages ($S > \SI{0}{\times}$).

\subsection{HyCASS CR Adapter Encoder Module}
After capturing spectral and spatial redundancies using the respective encoder modules, \ac{ours} applies the \ac{cr} adapter encoder module $E_\Psi: \mathbb{R}^{\Sigma \times \Omega \times N} \rightarrow \mathbb{R}^{\Sigma \times \Omega \times \Gamma}$ defined as:
\begin{align}
    E_\Psi \left( \matr{\xi} \right) = \text{Sigmoid} \left( \text{Conv2D}_{\qtyproduct{1x1}{}}^{N \rightarrow \Gamma} \left( \matr{\xi} \right) \right) .
\end{align}
It adjusts the number of latent channels $\Gamma$ to fit the targeted spatio-spectral \ac{cr}.
Therefore, a \qtyproduct{1x1}{} convolutional layer maps the spatio-spectral features from $N$ channels to $\Gamma$ latent channels.
We use the sigmoid activation function to constrain the latent space to the range \SIrange{0}{1}{}.

The spatio-spectral \ac{cr} achieved by our proposed model arises from the joint contributions of both spectral and spatial compression. It can be expressed as:
\begin{align}
    \ac{cr} = \ac{cr}_{\text{spec}} \cdot \ac{cr}_{\text{spat}} = \frac{C}{\Gamma} \cdot 4^{S},
    \label{eq:ours-cr}
\end{align}
where $\ac{cr}_{\text{spec}} = \frac{C}{\Gamma}$ denotes the spectral \ac{cr} determined by the reduction of spectral channels from $C$ to $\Gamma$, and $\ac{cr}_{\text{spat}} = 4^{S}$ corresponds to the spatial \ac{cr} introduced through $S$ stages of spatial downsampling.

\hl{Although the spectral encoder module and the {\ac{cr}} adapter encoder module are similarly structured, their roles differ: The spectral encoder module serves as an initial long-range spectral feature extractor, whereas the {\ac{cr}} adapter encoder module's main purpose is to adjust the number of latent channels.}

\subsection{HyCASS CR Adapter Decoder Module}
As the first step to perform reconstruction, the \ac{cr} adapter decoder module $D_{\Psi'}: \mathbb{R}^{\Sigma \times \Omega \times \Gamma} \rightarrow \mathbb{R}^{\Sigma \times \Omega \times N}$ of \ac{ours} defined as:
\begin{align}
    D_{\Psi'} \left( \matr{\xi} \right) = \text{\acs{leakyrelu}} \left( \text{Conv2D}_{\qtyproduct{1x1}{}}^{\Gamma \rightarrow N} \left( \matr{\xi} \right) \right)
\end{align}
applies a \qtyproduct{1x1}{} convolution that projects the channels from $\Gamma$ back to $N$, to match the \ac{ntu} dimension, followed by a \ac{leakyrelu}.

\subsection{HyCASS Spatial Decoder Module}
Afterwards, the spatial decoder module $D_{\chi'}: \mathbb{R}^{\Sigma \times \Omega \times N} \rightarrow \mathbb{R}^{H \times W \times N}$ of \ac{ours} defined as:
\begin{align}
    \begin{cases}
        D_\chi' \left( \matr{\xi} \right) = g^{\left( S \right)} \left( \matr{\xi} \right), \quad S \in \mathbb{N}_0 \\
        g  \left( \matr{\xi} \right) = \text{\acs{rstb}} \left( \text{\acs{leakyrelu}} \left( \text{Conv2D}_{\qtyproduct{3x3}{}}^{\uparrow^2} \left( \matr{\xi} \right) \right) \right)
    \end{cases}
\end{align}
applies $S$ stacked \acp{ntu} like the \ac{ours} spatial encoder module.
However, in each \ac{ntu}, first the \ac{2d} convolution is applied to aggregate short-range spatial features and upsample the spatial dimensions. Then, the \ac{rstb} is applied for long-range spatial feature extraction.

\subsection{HyCASS Spectral Decoder Module}
Finally, the \ac{ours} spectral decoder module $D_{\Phi'}: \mathbb{R}^{H \times W \times N} \rightarrow \mathbb{R}^{H \times W \times C}$ defined as:
\begin{align}
    D_{\Phi'} \left( \matr{\xi} \right) = \text{Sigmoid} \left( \text{Conv2D}_{\qtyproduct{1x1}{}}^{N \rightarrow C} \left( \matr{\xi} \right) \right)
\end{align}
projects the $N$ channels back to the original $C$ spectral bands using a \ac{2d} convolution with a \qtyproduct{1x1}{} kernel.
A sigmoid activation constrains the reconstructed output intensities to the valid range of \SIrange{0}{1}{}.

\section{Dataset Description and Experimental Setup}
\label{sec:dataset-description-and-experimental-setup}

\subsection{Dataset Description}
\hl{Three} \ac{hsi} datasets were employed in our experiments.
These datasets differ in several aspects, including spatial resolution, number of spectral bands, and dataset size, as summarized below.

\subsubsection{HySpecNet-11k}
\begin{figure*}
    \begin{minipage}[b]{.118\linewidth}
        \centering
        \includegraphics[width=\linewidth]{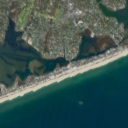}
    \end{minipage}
    \hfill
    \begin{minipage}[b]{.118\linewidth}
        \centering
        \includegraphics[width=\linewidth]{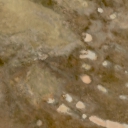}
    \end{minipage}
    \hfill
    \begin{minipage}[b]{.118\linewidth}
        \centering
        \includegraphics[width=\linewidth]{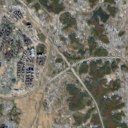}
    \end{minipage}
    \hfill
    \begin{minipage}[b]{.118\linewidth}
        \centering
        \includegraphics[width=\linewidth]{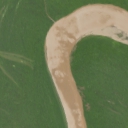}
    \end{minipage}
    \hfill
    \begin{minipage}[b]{.118\linewidth}
        \centering
        \includegraphics[width=\linewidth]{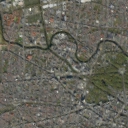}
    \end{minipage}
    \hfill
    \begin{minipage}[b]{.118\linewidth}
        \centering
        \includegraphics[width=\linewidth]{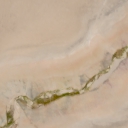}
    \end{minipage}
    \hfill
    \begin{minipage}[b]{.118\linewidth}
        \centering
        \includegraphics[width=\linewidth]{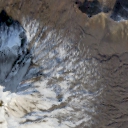}
    \end{minipage}
    \hfill
    \begin{minipage}[b]{.118\linewidth}
        \centering
        \includegraphics[width=\linewidth]{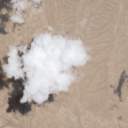}
    \end{minipage}
    \begin{minipage}[b]{.118\linewidth}
        \vspace{.055\linewidth}
        \centering
        \includegraphics[width=\linewidth]{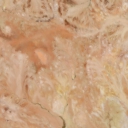}
    \end{minipage}
    \hfill
    \begin{minipage}[b]{.118\linewidth}
        \centering
        \includegraphics[width=\linewidth]{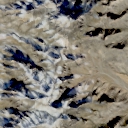}
    \end{minipage}
    \hfill
    \begin{minipage}[b]{.118\linewidth}
        \centering
        \includegraphics[width=\linewidth]{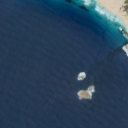}
    \end{minipage}
    \hfill
    \begin{minipage}[b]{.118\linewidth}
        \centering
        \includegraphics[width=\linewidth]{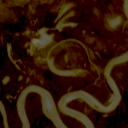}
    \end{minipage}
    \hfill
    \begin{minipage}[b]{.118\linewidth}
        \centering
        \includegraphics[width=\linewidth]{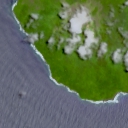}
    \end{minipage}
    \hfill
    \begin{minipage}[b]{.118\linewidth}
        \centering
        \includegraphics[width=\linewidth]{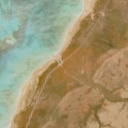}
    \end{minipage}
    \hfill
    \begin{minipage}[b]{.118\linewidth}
        \centering
        \includegraphics[width=\linewidth]{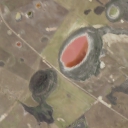}
    \end{minipage}
    \hfill
    \begin{minipage}[b]{.118\linewidth}
        \centering
        \includegraphics[width=\linewidth]{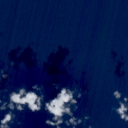}
    \end{minipage}
    \caption{An example of \acp{hsi} present in the HySpecNet-11k dataset \cite{fuchs2023hyspecnet}.}
    \label{fig:img-hyspecnet-11k}
\end{figure*}

HySpecNet-11k \cite{fuchs2023hyspecnet} is a large-scale hyperspectral benchmark dataset constructed from \num{250} tiles acquired by the \ac{enmap} satellite \cite{guanter2015enmap}.
It includes \num{11483} nonoverlapping \acp{hsi}, each of which consists of \qtyproduct{128x128}{} \si{\pixels} and \SI{202}{\sband} with a \ac{gsd} of \SI{30}{\meter} (low spatial resolution) and a spectral range of {\SIrange{420}{2450}{\nano\meter}}.
The data is radiometrically, geometrically, and atmospherically corrected (i.e., the \acs{l2a} water \& land product).
We used the recommended splits from \cite{fuchs2023hyspecnet} for training, validation, and test sets covering \SI{70}{\percent}, \SI{20}{\percent}, and \SI{10}{\percent} of the \acp{hsi}, respectively.
\autoref{fig:img-hyspecnet-11k} illustrates example images from this dataset.
We would like to note that we used HySpecNet-11k to show the effectiveness and generalization capability of our proposed model, particularly in the context of large-scale training with low spatial resolution \acp{hsi}.

\subsubsection{\hl{Berlin-Urban-Gradient}}
\begin{figure*}
    \begin{minipage}[b]{.07375\linewidth}
        \centering
        \includegraphics[width=\linewidth]{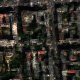}
    \end{minipage}
    \hfill
    \begin{minipage}[b]{.07375\linewidth}
        \centering
        \includegraphics[width=\linewidth]{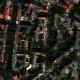}
    \end{minipage}
    \hfill
    \begin{minipage}[b]{.07375\linewidth}
        \centering
        \includegraphics[width=\linewidth]{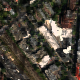}
    \end{minipage}
    \hfill
    \begin{minipage}[b]{.07375\linewidth}
        \centering
        \includegraphics[width=\linewidth]{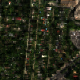}
    \end{minipage}
    \hfill
    \begin{minipage}[b]{.07375\linewidth}
        \centering
        \includegraphics[width=\linewidth]{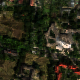}
    \end{minipage}
    \hfill
    \begin{minipage}[b]{.07375\linewidth}
        \centering
        \includegraphics[width=\linewidth]{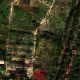}
    \end{minipage}
    \hfill
    \begin{minipage}[b]{.07375\linewidth}
        \centering
        \includegraphics[width=\linewidth]{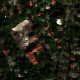}
    \end{minipage}
    \hfill
    \begin{minipage}[b]{.07375\linewidth}
        \centering
        \includegraphics[width=\linewidth]{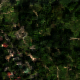}
    \end{minipage}
    \hfill
    \begin{minipage}[b]{.07375\linewidth}
        \centering
        \includegraphics[width=\linewidth]{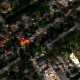}
    \end{minipage}
    \hfill
    \begin{minipage}[b]{.07375\linewidth}
        \centering
        \includegraphics[width=\linewidth]{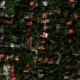}
    \end{minipage}
    \hfill
    \begin{minipage}[b]{.07375\linewidth}
        \centering
        \includegraphics[width=\linewidth]{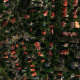}
    \end{minipage}
    \hfill
    \begin{minipage}[b]{.07375\linewidth}
        \centering
        \includegraphics[width=\linewidth]{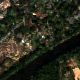}
    \end{minipage}
    \caption{\hl{An example of {\acp{hsi}} present in Berlin-Urban-Gradient dataset {\cite{okujeni2016berlin}}.}}
    \label{fig:img-berlin-urban-gradient}
\end{figure*}

\hl{Berlin-Urban-Gradient {\cite{okujeni2016berlin}} is a hyperspectral dataset covering a region along the urban-rural gradient of Berlin.
It comprises an airborne scene captured by the HyMap sensor {\cite{cocks1998hymaptm}} with {\SI{3.6}{\meter}} {\ac{gsd}} (medium spatial resolution) and a spectral range between {\SI{450}{\nano\meter}} and {\SI{2500}{\nano\meter}}, which was acquired on 20 August 2009 under clear sky conditions.
The tile was split into {\num{160}} non-overlapping {\acp{hsi}} of size ${\num{80}} \times {\num{80}}$ {\si{\pixels}} with {\SI{111}{\sband}} each.
We split the data into}
\begin{enumerate*}[i)]
    \item \hl{a training set that includes {\SI{70}{\percent}} of the {\acp{hsi}};}
    \item \hl{a validation set that includes {\SI{20}{\percent}} of the {\acp{hsi}}; and}
    \item \hl{test set that includes {\SI{10}{\percent}} of the {\acp{hsi}}.}
\end{enumerate*}
\hl{{\autoref{fig:img-berlin-urban-gradient}} shows exemplary images of this dataset.
We would like to note that we used the Berlin-Urban-Gradient dataset to show the effectiveness of {\ac{ours}} on {\acp{hsi}} with a medium spatial resolution and only few training samples.}

\subsubsection{MLRetSet}
\begin{figure*}
    \begin{minipage}[b]{.093\linewidth}
        \centering
        \includegraphics[width=\linewidth]{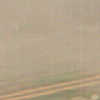}
    \end{minipage}
    \hfill
    \begin{minipage}[b]{.093\linewidth}
        \centering
        \includegraphics[width=\linewidth]{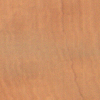}
    \end{minipage}
    \hfill
    \begin{minipage}[b]{.093\linewidth}
        \centering
        \includegraphics[width=\linewidth]{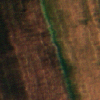}
    \end{minipage}
    \hfill
    \begin{minipage}[b]{.093\linewidth}
        \centering
        \includegraphics[width=\linewidth]{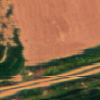}
    \end{minipage}
    \hfill
    \begin{minipage}[b]{.093\linewidth}
        \centering
        \includegraphics[width=\linewidth]{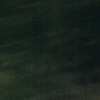}
    \end{minipage}
    \hfill
    \begin{minipage}[b]{.093\linewidth}
        \centering
        \includegraphics[width=\linewidth]{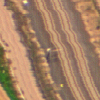}
    \end{minipage}
    \hfill
    \begin{minipage}[b]{.093\linewidth}
        \centering
        \includegraphics[width=\linewidth]{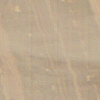}
    \end{minipage}
    \hfill
    \begin{minipage}[b]{.093\linewidth}
        \centering
        \includegraphics[width=\linewidth]{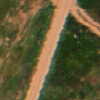}
    \end{minipage}
    \hfill
    \begin{minipage}[b]{.093\linewidth}
        \centering
        \includegraphics[width=\linewidth]{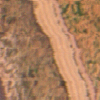}
    \end{minipage}
    \hfill
    \begin{minipage}[b]{.093\linewidth}
        \centering
        \includegraphics[width=\linewidth]{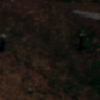}
    \end{minipage}
    \begin{minipage}[b]{.093\linewidth}
        \vspace{.07\linewidth}
        \centering
        \includegraphics[width=\linewidth]{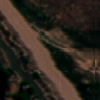}
    \end{minipage}
    \hfill
    \begin{minipage}[b]{.093\linewidth}
        \centering
        \includegraphics[width=\linewidth]{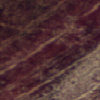}
    \end{minipage}
    \hfill
    \begin{minipage}[b]{.093\linewidth}
        \centering
        \includegraphics[width=\linewidth]{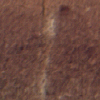}
    \end{minipage}
    \hfill
    \begin{minipage}[b]{.093\linewidth}
        \centering
        \includegraphics[width=\linewidth]{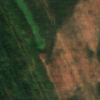}
    \end{minipage}
    \hfill
    \begin{minipage}[b]{.093\linewidth}
        \centering
        \includegraphics[width=\linewidth]{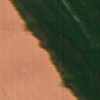}
    \end{minipage}
    \hfill
    \begin{minipage}[b]{.093\linewidth}
        \centering
        \includegraphics[width=\linewidth]{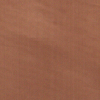}
    \end{minipage}
    \hfill
    \begin{minipage}[b]{.093\linewidth}
        \centering
        \includegraphics[width=\linewidth]{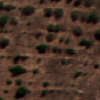}
    \end{minipage}
    \hfill
    \begin{minipage}[b]{.093\linewidth}
        \centering
        \includegraphics[width=\linewidth]{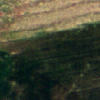}
    \end{minipage}
    \hfill
    \begin{minipage}[b]{.093\linewidth}
        \centering
        \includegraphics[width=\linewidth]{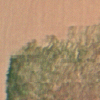}
    \end{minipage}
    \hfill
    \begin{minipage}[b]{.093\linewidth}
        \centering
        \includegraphics[width=\linewidth]{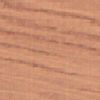}
    \end{minipage}
    \caption{An example of \acp{hsi} present in the MLRetSet dataset \cite{omruuzun2024novel}.}
    \label{fig:img-mlretset}
\end{figure*}

MLRetSet \cite{omruuzun2024novel} is a hyperspectral benchmark dataset created from high spatial resolution hyperspectral imagery with \SI{27.86}{\centi\meter} \ac{gsd}.
The hyperspectral dataset was acquired during an airborne flight covering the Turkish towns Yenice and Yeşilkaya on 4 May 2019.
The twelve acquired tiles were split into \num{3840} non-overlapping \acp{hsi} of size \qtyproduct{100x100}{} \si{\pixels} with \SI{369}{\sband} each.
We split the data into
\begin{enumerate*}[i)]
    \item a training set that includes \SI{70}{\percent} of the \acp{hsi}; 
    \item a validation set that includes \SI{20}{\percent} of the \acp{hsi}; and
    \item test set that includes \SI{10}{\percent} of the \acp{hsi}.
\end{enumerate*}
\autoref{fig:img-mlretset} provides visual examples of typical scenes present in the MLRetSet dataset.
We would like to note that we employed the MLRetSet dataset to demonstrate the effectiveness of our proposed model on \acp{hsi} with a high spatial resolution.

\subsection{Experimental Setup}
Our code was implemented in PyTorch based on the CompressAI \cite{begaint2020compressai} framework.
In the case of MLRetSet, the \acp{hsi} were center-cropped to \qtyproduct{96x96}{} \si{\pixels} to facilitate repeated spatial downsampling by factors of two, which is required for spatial compression models.
For HySpecNet-11k, we followed the easy split as introduced in \cite{fuchs2023hyspecnet}.
Training runs were carried out on a single NVIDIA A100 SXM4 80 GB \acs{gpu} using the Adam optimizer \cite{kingma2014adam}.
For the loss function, we employed the \acf{mse} \hl{between the original {\ac{hsi}} $\vect{X} \in \mathbb{R}^{H \times W \times C}$ and its reconstruction $\vect{\hat{X}} \in \mathbb{R}^{H \times W \times C}$}:
\begin{align}
    \text{MSE}(\vect{X}, \vect{\hat{X}}) = \frac{1}{H \cdot W \cdot C} \sum_{h,w,c} \left( \vect{X}(h,w,c) - \vect{\hat{X}}(h,w,c) \right)^2.
    \label{eq:mse}
\end{align}
\hl{Here, $H$, $W$ and $C$ denote {\ac{hsi}} height, width and spectral bands, respectively while $\vect{X}(h,w,c)$ represents the value of spectral band $c$ at pixel location $(h,w)$.}

We compare our proposed model with two traditional compression methods (JPEG2000 and \ac{pca}) and the following learning-based models:
\begin{enumerate*}[1)]
    \item \acs{1dcae} \cite{kuester20211d}, a \ac{1d} convolutional \ac{ae} performing spectral compression;
    \item \acs{sscnet} \cite{la2022hyperspectral}, a \ac{2d} convolutional \ac{ae} performing spatial compression;
    \item \acs{3dcae} \cite{chong2021end}, a \ac{3d} convolutional \ac{ae} jointly compressing spatial and spectral redundancies;
    \item \hl{Verdú et al. {\cite{mijares2023scalable}}, a {\ac{2d}} variational {\ac{ae}} based on channel clustering}; and
    \item \acs{hycot} \cite{fuchs2024hycot}, a transformer-based \ac{ae} exploiting spectral redundancies.  
\end{enumerate*}
Training epochs, \ac{lr}, and \ac{bs} were adjusted per model and dataset to optimize training time and \acs{gpu} memory usage, while ensuring convergence of the loss function for each training run.
\hl{To this end, the optimal hyperparameters were empirically determined through preliminary experiments (grid search) and analysis of the obtained results.}
\autoref{tab:hyperparameters} lists the specific training hyperparameters used for each model and dataset configuration.
It is worth noting that on HySpecNet-11k for \ac{1dcae}, the number of epochs was reduced to \num{250} for \ac{cr} $\in \left\{ \num{8}, \num{16}, \num{32} \right\}$ due to runtime limitations.
For MLRetSet, we had to reduce the \ac{bs} to \num{1} for \ac{1dcae} at $\ac{cr} = \num{32}$ due to \acs{gpu} memory constraints.
Furthermore, we used the random sampling strategy from \cite{fuchs2024hycot} to efficiently train \ac{hycot}.

For \ac{ours}, we fixed the number of spatial encoder module channels to $N = \num{128}$ and the \acs{swin} Transformer's window size was set to \num{8}, consistent with the configuration used in the \ac{tic} model \cite{lu2021transformer}, and all remaining parameters were aligned accordingly.
The number of spatial stages was varied between \SI{0}{\times} and \SI{3}{\times} to assess the effect of different spatial compression levels.
We would like to note that for zero spatial stages, we increased $N$ to \num{1024} to compensate for the missing spatial stages in terms of model parameters and \ac{flops}.
Also, we applied the random sampling strategy from \ac{hycot} \cite{fuchs2024hycot} and increased the \ac{lr} to \num{1e-3}, the \ac{bs} to \num{64}, and the number of epochs to \num{2000} for HySpecNet-11k \hl{and Berlin-Urban-Gradient} and \num{1000} for MLRetSet.
\hl{For the Berlin-Urban-Gradient dataset, we had to reduce the window size to {\num{4}} when using {\SI{3}{\times}} spatial stages.}
It is important to note that while the targeted spatio-spectral \acp{cr} were $\ac{cr} \in \left\{ \num{4}, \num{8}, \num{16}, \dots, 1024 \right\}$, the achieved \acp{cr} may deviate due to the spectral dimension not being divisible by powers of two.

\begin{table*}
    \centering
    \caption{Values of training hyperparameter selected for each model for the HySpecNet-11k \cite{fuchs2023hyspecnet}, \hl{Berlin-Urban-Gradient {\cite{okujeni2016berlin}}} and MLRetSet \cite{omruuzun2024novel} dataset.}
    \label{tab:hyperparameters}
    \begin{tabular}{c S[table-format=4.0] S[table-format=0.0e-1] S[table-format=2.0] S[table-format=4.0] S[table-format=0.0e-1] S[table-format=2.0] S[table-format=3.0] S[table-format=0.0e-1] S[table-format=2.0]}
        \hline
        \multirow{2}{*}{\vspace{-0.6em}Model} & \multicolumn{3}{c}{HySpecNet-11k \cite{fuchs2023hyspecnet}} & \multicolumn{3}{c}{\hl{Berlin-Urban-Gradient {\cite{okujeni2016berlin}}}} & \multicolumn{3}{c}{MLRetSet \cite{omruuzun2024novel}} \\
        \cmidrule(lr){2-4} \cmidrule(lr){5-7} \cmidrule(lr){8-10}
         & {Epochs} & \acs{lr} & \acs{bs} & \hl{{Epochs}} & \hl{{\acs{lr}}} & \hl{{\acs{bs}}} & {Epochs} & \acs{lr} & \acs{bs} \\
         \hline
        \acs{1dcae} \cite{kuester20211d} & 500 & 1e-4 & 2 & 500 & 1e-4 & 2 & 100 & 1e-4 & 2 \\
        \acs{sscnet} \cite{la2022hyperspectral} & 2000 & 1e-5 & 8 & 2000 & 1e-5 & 8 & 400 & 1e-5 & 8 \\
        \acs{3dcae} \cite{chong2021end} & 1000 & 1e-4 & 2 & 1000 & 1e-4 & 2 & 250 & 1e-4 & 2 \\
        \acs{hycot} \cite{fuchs2024hycot} & 2000 & 1e-3 & 64 & 2000 & 1e-3 & 64 & 500 & 1e-3 & 64 \\
        \acs{ours} & 200 & 1e-4 & 16 & 1000 & 1e-4 & 16 & 100 & 1e-4 & 16 \\
    \end{tabular}
\end{table*}

\subsection{Evaluation Metrics}
For the evaluation of the \ac{hsi} compression methods,  we consider two kinds of metrics:
\begin{itemize*}[i)]
    \item metrics that measure the compression efficiency; and
    \item metrics that measure the reconstruction quality.
\end{itemize*}
In our experiments, we use the \ac{cr} to quantify the data reduction.
\Ac{psnr}, \ac{sa} and \ac{ssim} are used to measure the fidelity of a reconstructed \ac{hsi}.
Given an original \ac{hsi} $\vect{X} \in \mathbb{R}^{H \times W \times C}$, its latent representation $\vect{Y} \in \mathbb{R}^{\Sigma \times \Omega \times \Gamma}$ and reconstruction $\vect{\hat{X}} \in \mathbb{R}^{H \times W \times C}$, the metrics are defined as follows.

\subsubsection{Compression Ratio (CR)}
The \ac{cr} between an original \ac{hsi} $\vect{X}$ with bit depth $N_b$ and its representation in the latent space $\vect{Y}$ with bit depth $\hat{N}_b$ after encoding can be expressed as follows:
\begin{align}
    \text{\acs{cr}} \left( \vect{X}, \vect{Y} \right) = \frac{N_b \cdot H \cdot W \cdot C}{\hat{N}_b \cdot \Sigma \cdot \Omega \cdot \Gamma} .
\end{align}
A higher \ac{cr} indicates greater compression of the hyperspectral data, which could result in increased loss of information during reconstruction.

\subsubsection{Peak Signal-to-Noise Ratio (PSNR)}
For measuring the reconstruction quality, we use the \ac{psnr} between original \ac{hsi} $\vect{X}$ and reconstructed \ac{hsi} $\vect{\hat{X}}$, which is defined as:
\begin{align}
    \text{\acs{psnr}} \left( \vect{X}, \vect{\hat{X}} \right) = 10 \cdot \log_{10} \left( \frac{\text{MAX}^2}{\text{MSE} \left( \vect{X}, \vect{\hat{X}} \right)} \right),
\end{align}
where $\text{MAX}$ denotes the maximum possible pixel value (e.g. \num{1.0} in the case of min-max normalization), and the \ac{mse} is defined as in \autoref{eq:mse}.
A higher \ac{psnr} value indicates better reconstruction quality with less distortion.

\subsubsection{Spectral Angle (SA)}
We also report the \ac{sa} defined as:
\begin{equation}
\begin{aligned}
    &\text{\acs{sa}} \left( \vect{X}, \vect{\hat{X}} \right) = \frac{1}{H \cdot W} \\
    & \sum_{h, w} \frac{180}{\pi} \arccos \left( \frac{\sum_c \vect{X} \left( h, w, c \right) \cdot \vect{\hat{X}} \left( h, w, c \right)}{\sqrt{\sum_c \vect{X} \left( h, w, c \right)^2} \sqrt{\sum_c \vect{\hat{X}} \left( h, w, c \right)^2}} \right),
\end{aligned}
\end{equation}
which quantifies the average spectral similarity between all pixels of an original $\vect{X}$ and a reconstructed \ac{hsi} $\vect{\hat{X}}$.
A smaller \ac{sa} indicates higher spectral similarity and is inherently scale-invariant.

\subsubsection{\hl{Structural Similarity Index Measure (SSIM)}}
\hl{To measure the perceived quality of a reconstruction, we use the {\ac{ssim}} defined as:}
\begin{align}
    \text{\acs{ssim}} \left( \vect{X}, \vect{\hat{X}} \right) = 
    \frac{(2\mu_X \mu_{\hat{X}} + c_1)(2\sigma_{X\hat{X}} + c_2)}{(\mu_X^2 + \mu_{\hat{X}}^2 + c_1)(\sigma_X^2 + \sigma_{\hat{X}}^2 + c_2)}.
\end{align}
\hl{Where $\mu_X$ and $\mu_{\hat{X}}$ are the means, $\sigma_X^2$ and $\sigma_{\hat{X}}^2$ are the variances and $\sigma_{X\hat{X}}$ is the covariance of $\vect{X}$ and $\vect{\hat{X}}$. $c_1$ and $c_2$ are constants to provide stability against weak denominators. The {\ac{ssim}} metric yields values ranging from {\num{-1}} to {\num{1}}, with higher values indicating reconstructions of higher perceptual quality.}

\section{Experimental Results}
\label{sec:experimental-results}
We conducted three sets of experiments, aiming at:
\begin{enumerate*}[1)]
    \item assessment of the effects of spectral and spatial compression within the proposed \ac{ours} model through an ablation study on \hl{three} benchmark datasets;
    \item comparison of our model's effectiveness with traditional baselines and lossy learning-based state-of-the-art \ac{hsi} compression models; and
    \item qualitative analysis of the reconstruction results.
\end{enumerate*}

\subsection{Ablation Study}
\label{subsec:experimenta-results-ablation-study}
In this subsection, we analyze the impact of varying the spatial stages $S$ and latent channels $\Gamma$ inside \ac{ours} on the reconstruction quality for multiple \acp{cr}.
To assess generalization across different spatial resolutions, we report results on the HySpecNet-11k\hl{, the Berlin-Urban-Gradient} and the MLRetSet datasets.
For each configuration, $S$ defines the spatial \ac{cr} ($\ac{cr}_\text{spat}$), while $\Gamma$ determines the spectral \ac{cr} ($\ac{cr}_\text{spec}$).
Given $\ac{cr}_\text{spat}$, $\ac{cr}_\text{spec}$ is adjusted accordingly to match the targeted spatio-spectral \ac{cr}.

\subsubsection{HySpecNet-11k}
\autoref{tab:results-ablation-hyspecnet11k} shows the results of the ablation study on the HySpecNet-11k dataset.
\begin{table}
    \centering
    \caption{\ac{ours} results obtained by varying the spatial stages $S$ on the easy split test set of the HySpecNet-11k \cite{fuchs2023hyspecnet} dataset. $\ac{cr}_\text{spec}$, $\ac{cr}_\text{spat}$ and \ac{cr} denote spectral, spatial and joint spatio-spectral \acl{cr}, respectively. Reconstruction quality is evaluated using \ac{psnr} and \ac{sa}.}
    \label{tab:results-ablation-hyspecnet11k}
    \begin{tabular}{S S[round-mode=places,round-precision=2,detect-weight=true] S[round-mode=places,round-precision=2,detect-weight=true] S[round-mode=places,round-precision=2,detect-weight=true] H S[round-mode=places,round-precision=2,detect-weight=true] S[round-mode=places,round-precision=2,detect-weight=true] H H}
        \hline
        $S$ & \ac{cr} & $\ac{cr}_\text{spec}$ & $\ac{cr}_\text{spat}$ & {Latent Space} & \ac{psnr} $\uparrow$ & \ac{sa} $\downarrow$ & \acs{flops} $\downarrow$ & Parameters $\downarrow$ \\
        \hline
        \SI{0}{\times} & 3.9608 & 3.9608 & 1.0 & \qtyproduct{128x128x51}{} & \textbf{\SI{56.444}{\decibel}} & \textbf{\ang{1.3940}} & \SI{17.06}{\giga\nothing} & 520445 \\          
        \SI{1}{\times} & 3.9608 & 0.9902 & 4.0 & \qtyproduct{64x64x204}{} & \SI{49.779}{\decibel} & \ang{2.4409} & \SI{8.32}{\giga\nothing} & 1196454 \\
        \SI{2}{\times} & 3.9656 & 0.24785 & 16.0 & \qtyproduct{32x32x815}{} & \SI{48.447}{\decibel} & \ang{2.6473} & \SI{9.90}{\giga\nothing} & 3249225 \\
        \SI{3}{\times} & 3.9669 & 0.061982812 & 64.0 & \qtyproduct{16x16x3259}{} & \SI{44.647}{\decibel} & \ang{3.3339} & \SI{10.30}{\giga\nothing} & 6594965 \\
        \hline
        \SI{0}{\times} & 7.7692 & 7.692 & 1.0 & \qtyproduct{128x128x26}{} & \textbf{\SI{55.155}{\decibel}} & \textbf{\ang{1.5574}} & \SI{14.5}{\giga\nothing} & 442583 \\
        \SI{1}{\times} & 7.7692 & 1.9423 & 4.0 & \qtyproduct{64x64x104}{} & \SI{49.832}{\decibel} & \ang{2.4431} & \SI{8.10}{\giga\nothing} & 1170754 \\
        \SI{2}{\times} & 7.7692 & 0.485575 & 16.0 & \qtyproduct{32x32x416}{} & \SI{48.084}{\decibel} & \ang{2.7065} & \SI{9.68}{\giga\nothing} & 3146682 \\
        \SI{3}{\times} & 7.7692 & 0.121 & 64.0 & \qtyproduct{16x16x1664}{} & \SI{44.788}{\decibel} & \ang{3.2651} & \SI{10.1}{\giga\nothing} & 6185050 \\
        \hline
        \SI{0}{\times} & 15.538 & 15.538 & 1.0 & \qtyproduct{128x128x13}{} & \textbf{\SI{52.828}{\decibel}} & \textbf{\ang{1.8364}} & \SI{14.5}{\giga\nothing} & 442583 \\        
        \SI{1}{\times} & 15.538 & 3.8845 & 4.0 & \qtyproduct{64x64x52}{} & \SI{50.392}{\decibel} & \ang{2.3918} & \SI{8.00}{\giga\nothing} & 1157390 \\
        \SI{2}{\times} & 15.538 & 0.971125 & 16.0 & \qtyproduct{32x32x208}{} & \SI{48.798}{\decibel} & \ang{2.6387} & \SI{9.58}{\giga\nothing} & 3093226 \\
        \SI{3}{\times} & 15.538 & 0.24278125 & 64.0 & \qtyproduct{16x16x832}{} & \SI{46.057}{\decibel} & \ang{3.0827} & \SI{9.98}{\giga\nothing} & 5971226 \\
        \hline
        \SI{0}{\times} & 28.857 & 28.857 & 1.0 & \qtyproduct{128x128x7}{} & \textbf{\SI{49.719}{\decibel}} & \textbf{\ang{2.2678}} & \SI{14.10}{\giga\nothing} & 430289 \\
        \SI{1}{\times} & 28.857 & 7.21425 & 4.0 & \qtyproduct{64x64x28}{} & \SI{49.249}{\decibel} & \ang{2.5456} & \SI{7.94}{\giga\nothing} & 1151222 \\
        \SI{2}{\times} & 28.857 & 1.8035625 & 16.0 & \qtyproduct{32x32x112}{} & \SI{48.031}{\decibel} & \ang{2.7195} & \SI{9.52}{\giga\nothing} & 3068554 \\
        \SI{3}{\times} & 28.857 & 0.450890625 & 64.0 & \qtyproduct{16x16x448}{} & \SI{44.940}{\decibel} & \ang{3.2424} & \SI{9.94}{\giga\nothing} & 5872538 \\
        \hline
        \SI{0}{\times} & 50.50 & 50.50 & 1.0 & \qtyproduct{128x128x4}{} & \SI{45.862}{\decibel} & \ang{3.1516} & \SI{13.90}{\giga\nothing} & 424142 \\
        \SI{1}{\times} & 50.50 & 12.625 & 4.0 & \qtyproduct{64x64x16}{} & \textbf{\SI{48.610}{\decibel}} & \ang{2.7177} & \SI{7.92}{\giga\nothing} & 1148138 \\
        \SI{2}{\times} & 50.50 & 3.15625 & 16.0 & \qtyproduct{32x32x64}{} & \SI{48.579}{\decibel} & \textbf{\ang{2.6440}} & \SI{9.50}{\giga\nothing} & 3056218 \\
        \SI{3}{\times} & 50.50 & 0.7890625 & 64.0 & \qtyproduct{16x16x256}{} & \SI{45.916}{\decibel} & \ang{3.1105} & \SI{9.90}{\giga\nothing} & 5823194 \\
        \hline
        \SI{0}{\times} & 101.00 & 101.00 & 1.0 & \qtyproduct{128x128x2}{} & \SI{39.836}{\decibel} & \ang{5.5225} & \SI{13.76}{\giga\nothing} & 420044 \\
        \SI{1}{\times} & 101.00 & 25.25 & 4.0 & \qtyproduct{64x64x8}{} & \SI{45.969}{\decibel} & \ang{3.155} & \SI{7.90}{\giga\nothing} & 1146082 \\
        \SI{2}{\times} & 101.00 & 6.3125 & 16.0 & \qtyproduct{32x32x32}{} & \textbf{\SI{46.843}{\decibel}} & \textbf{\ang{2.9066}} & \SI{9.48}{\giga\nothing} & 3047994 \\
        \SI{3}{\times} & 101.00 & 1.578125 & 64.0 & \qtyproduct{16x16x128}{} & \SI{44.441}{\decibel} & \ang{3.3200} & \SI{9.90}{\giga\nothing} & 5790298 \\
        \hline
        \SI{0}{\times} & 202.00 & 202.00 & 1.0 & \qtyproduct{128x128x1}{} & \SI{32.971}{\decibel} & \ang{12.513} & \SI{13.70}{\giga\nothing} & 417995 \\
        \SI{1}{\times} & 202.00 & 50.5 & 4.0 & \qtyproduct{64x64x4}{} & \SI{43.347}{\decibel} & \ang{3.7722} & \SI{7.90}{\giga\nothing} & 1145054 \\
        \SI{2}{\times} & 202.00 & 12.625 & 16.0 & \qtyproduct{32x32x16}{} & \textbf{\SI{45.094}{\decibel}} & \textbf{\ang{3.1558}} & \SI{9.48}{\giga\nothing} & 3043882 \\
        \SI{3}{\times} & 202.00 & 3.15625 & 64.0 & \qtyproduct{16x16x64}{} & \SI{44.136 }{\decibel} & \ang{3.4310} & \SI{9.88}{\giga\nothing} & 5773850 \\
        \hline
        \SI{1}{\times} & 404.00 & 101.0 & 4.0 & \qtyproduct{64x64x2}{} & \SI{41.051}{\decibel} & \ang{4.5494} & \SI{7.90}{\giga\nothing} & 1144540 \\
        \SI{2}{\times} & 404.00 & 25.25 & 16.0 & \qtyproduct{32x32x8}{} & \SI{42.652}{\decibel} & \ang{3.7228} & \SI{9.48}{\giga\nothing} & 3041826 \\
        \SI{3}{\times} & 404.00 & 6.3125 & 64.0 & \qtyproduct{16x16x32}{} & \textbf{\SI{42.758}{\decibel}} & \textbf{\ang{3.6519}} & \SI{9.88}{\giga\nothing} & 5765626 \\
        \hline
        \SI{1}{\times} & 808.00 & 202 & 4.0 & \qtyproduct{64x64x1}{} & \SI{36.385}{\decibel} & \ang{7.4544} & \SI{7.88}{\giga\nothing} & 1144283 \\
        \SI{2}{\times} & 808.00 & 50.5 & 16.0 & \qtyproduct{32x32x4}{} & \SI{40.814}{\decibel} & \ang{4.2197} & \SI{9.48}{\giga\nothing} & 3040798 \\
        \SI{3}{\times} & 808.00 & 12.625 & 64.0 & \qtyproduct{16x16x16}{} & \textbf{\SI{41.455}{\decibel}} & \textbf{\ang{3.9490}} & \SI{9.88}{\giga\nothing} & 5761514 \\
        \hline
    \end{tabular}
\end{table}
From the table, one can derive three key observations:

First, for $\acp{cr} < \num{32}$, \ac{ours} with zero spatial stages (i.e., spectral compression only) yields superior reconstruction performance compared to \ac{ours} with one or more spatial stages (i.e., spatio-spectral compression).
For example, when $\ac{cr} \approx \num{4}$, \ac{ours} with zero spatial stages achieves a \ac{psnr} of \SI{56.44}{\decibel} while reconstruction quality reduces for one, two and three spatial stages to \SI{49.78}{\decibel}, \SI{48.45}{\decibel} and \SI{44.65}{\decibel}, respectively.
This can be explained by the relatively high number of latent channels retained in this \ac{cr} range by \ac{ours} models without spatial compression (e.g. $\Gamma = \num{51}$ for $\ac{cr} \approx \num{4}$), which provide sufficient spectral information for accurate reconstruction by the decoder without requiring any spatial feature aggregation.
For HySpecNet-11k, spatial compression poses reconstruction challenges due to the limited spatial correlation caused by the low spatial resolution, leading to noticeably blurred reconstruction results.
Interestingly, at low \acp{cr}, spatio-spectral \ac{ours} models tend to exhibit \ac{psnr} saturation.
This behavior suggests that the high number of latent channels in such configurations contains redundancy, allowing comparable reconstruction performance at significantly higher \acp{cr}.

Second, as the \ac{cr} increases beyond \num{32}, the performance of \ac{ours} models that rely solely on spectral compression with zero spatial stages diminishes rapidly.
In contrast, \ac{ours} models that incorporate spatial compression (one or more spatial stages) maintain a higher quality of reconstruction at these compression levels.
This behavior indicates that with higher \acp{cr}, where spectral compression reaches saturation, the inclusion of deeper spatial hierarchies becomes increasingly important for preserving structural and spectral fidelity.
In particular, this trend persists even for $\acp{cr} > \num{64}$, where models with a higher number of spatial stages consistently achieve better reconstruction performance.
These findings highlight the increasing importance of spatial compression in highly constrained compression scenarios.

Third, reconstruction quality generally decreases with increasing \acp{cr}, dropping from \SI{56.44}{\decibel} at $\ac{cr} = \num{3.96}$ to \SI{41.46}{\decibel} at $\ac{cr} = \num{808}$ with the respective optimal spatial stages.
This trend is consistently observed by increasing \ac{sa} values, indicating that stronger compression leads to greater loss of spatial and spectral fidelity.

\subsubsection{\hl{Berlin-Urban-Gradient}}
\hl{\mbox{\autoref{tab:results-ablation-berlinurbangradient}} shows the results of the ablation study of {\ac{ours}} on the Berlin-Urban-Gradient dataset.}
\begin{table}
    \centering
    \caption{\hl{{\ac{ours}} results obtained by varying the spatial stages $S$ on the test set of the Berlin-Urban-Gradient {\cite{okujeni2016berlin}} dataset. ${\ac{cr}}_\text{spec}$, ${\ac{cr}}_\text{spat}$ and {\ac{cr}} denote spectral, spatial and joint spatio-spectral compression ratio, respectively. Reconstruction quality is evaluated using {\ac{psnr}} and {\ac{sa}}.}}
    \label{tab:results-ablation-berlinurbangradient}
    \begin{tabular}{S S[round-mode=places,round-precision=2,detect-weight=true] S[round-mode=places,round-precision=2,detect-weight=true] S[round-mode=places,round-precision=2,detect-weight=true] S[round-mode=places,round-precision=2,detect-weight=true] S[round-mode=places,round-precision=2,detect-weight=true]}
        \hline
        $S$ & \ac{cr} & $\ac{cr}_\text{spec}$ & $\ac{cr}_\text{spat}$ & \ac{psnr} $\uparrow$ & \ac{sa} $\downarrow$ \\
        \hline
        \SI{0}{\times} & 4.1111 & 4.1111 & 1 & \textbf{\SI{49.547}{\decibel}} & \textbf{\ang{1.1643}} \\
        \SI{1}{\times} & 4.1111 & 1.02775 & 4.0 & \SI{41.987}{\decibel} & \ang{2.3889} \\
        \SI{2}{\times} & 4.1111 & 0.25694375 & 16.0 & \SI{38.666}{\decibel} & \ang{3.4422} \\
        \SI{3}{\times} & 4.1111 & 0.064235938 & 64.0 & \SI{29.237}{\decibel} & \ang{6.4842} \\
        \hline
        \SI{0}{\times} & 7.9286 & 7.9286 & 1.0 & \textbf{\SI{49.499}{\decibel}} & \textbf{\ang{1.1773}} \\
        \SI{1}{\times} & 7.9286 & 1.98215 & 4.0 & \SI{41.775}{\decibel} & \ang{2.4538} \\
        \SI{2}{\times} & 7.9286 & 0.4955375 & 16.0 & \SI{37.797}{\decibel} & \ang{3.7782} \\
        \SI{3}{\times} & 7.9286 & 0.123884375 & 64.0 & \SI{25.502}{\decibel} & \ang{10.451} \\
        \hline
        \SI{0}{\times} & 15.857 & 15.857 & 1.0 & \textbf{\SI{47.534}{\decibel}} & \textbf{\ang{1.4520}} \\
        \SI{1}{\times} & 15.857 & 3.96425 & 4.0 & \SI{41.839}{\decibel} & \ang{2.4209} \\
        \SI{2}{\times} & 15.857 & 0.9736875 & 16.0 & \SI{37.337}{\decibel} & \ang{3.9539} \\
        \SI{3}{\times} & 15.857 & 0.247765625 & 64.0 & \SI{30.704}{\decibel} & \ang{5.4393} \\
        \hline
        \SI{0}{\times} & 37.000 & 37.000 & 1.0 & \textbf{\SI{43.684}{\decibel}} & \textbf{\ang{1.9527}} \\
        \SI{1}{\times} & 37.000 & 9.25 & 4.0 & \SI{39.655}{\decibel} & \ang{3.0302} \\
        \SI{2}{\times} & 37.000 & 2.3125 & 16.0 & \SI{37.525}{\decibel} & \ang{3.8412} \\
        \SI{3}{\times} & 37.000 & 0.578125 & 64.0 & \SI{29.781}{\decibel} & \ang{6.1963} \\
        \hline
        \SI{0}{\times} & 55.500 & 55.500 & 1.0 & \textbf{\SI{38.768}{\decibel}} & \textbf{\ang{3.0926}} \\
        \SI{1}{\times} & 55.500 & 13.875 & 4.0 & \SI{37.705}{\decibel} & \ang{3.8063} \\
        \SI{2}{\times} & 55.500 & 3.4375 & 16.00 & \SI{36.909}{\decibel} & \ang{3.8664} \\
        \SI{3}{\times} & 55.500 & 0.859375 & 64.00 & \SI{31.154}{\decibel} & \ang{5.4162} \\
        \hline
        \SI{0}{\times} & 111.00 & 111.00 & 1.0 & \SI{29.325}{\decibel} & \ang{8.8831} \\
        \SI{1}{\times} & 111.0 & 27.75 & 4.0 & \SI{34.982}{\decibel} & \ang{4.5481} \\
        \SI{2}{\times} & 111.0 & 6.9375 & 16.0 & \textbf{\SI{35.225}{\decibel}} & \textbf{\ang{4.3106}} \\
        \SI{3}{\times} & 111.0 & 1.734375 & 64.0 & \SI{25.735}{\decibel} & \ang{11.006} \\
        \hline
        \SI{1}{\times} & 222.0 & 55.5 & 4.0 & \SI{30.806}{\decibel} & \ang{5.8172} \\
        \SI{2}{\times} & 222.0 & 13.875 & 16.0 & \textbf{\SI{31.345}{\decibel}} & \textbf{\ang{5.3811}} \\
        \SI{3}{\times} & 222.0 & 3.46875 & 64.0 & \SI{29.931}{\decibel} & \ang{6.1617} \\
        \hline
        \SI{1}{\times} & 444.0 & 111.0 & 4.0 & \SI{26.135}{\decibel} & \ang{11.625} \\
        \SI{2}{\times} & 444.0 & 27.75 & 16.0 & \textbf{\SI{28.484}{\decibel}} & \textbf{\ang{6.4999}} \\
        \SI{3}{\times} & 444.0 & 6.9375 & 64.0 & \SI{25.555}{\decibel} & \ang{10.983} \\
        \hline
        \SI{2}{\times} & 888.0 & 55.5 & 16.0 & \textbf{\SI{26.194}{\decibel}} & \textbf{\ang{8.6331}} \\
        \SI{3}{\times} & 888.0 & 13.875 & 64.0 & \SI{26.141}{\decibel} & \ang{8.7347} \\
        \hline
        \SI{2}{\times} & 1776.0 & 111.0 & 16.0 & \textbf{\SI{24.519}{\decibel}} & \textbf{\ang{10.073}} \\
        \SI{3}{\times} & 1776.0 & 27.75 & 64.0 & \SI{24.494}{\decibel} & \ang{10.230} \\
        \hline
    \end{tabular}
\end{table}
\hl{The results in the table indicate that for ${\acp{cr}} < {\num{64}}$, purely spectral compression with {\SI{0}{\times}} spatial stages outperforms configurations that exploit spatial redundancies.
This is due to the available spectral redundancies inside the {\SI{111}{\sband}}, which for low {\acp{cr}} yield better compression potential than spatial compression through spatial stages.
Once the {\ac{cr}} increases further, limiting the latent space to only one latent channel, the spatial stages become essential to enable effective spatio-spectral compression.
Thus, for ${\acp{cr}} > {\num{64}}$, employing two spatial stages proves advantageous compared to spectral compression.
The immediate increase to two spatial stages highlights the importance of exploiting spatial redundancies in {\acp{hsi}} with medium spatial resolution at medium and high {\acp{cr}}.}

\subsubsection{MLRetSet}
\autoref{tab:results-ablation-mlretset} shows the ablation study of \ac{ours} on the MLRetSet dataset.
\begin{table}
    \centering
    \caption{\ac{ours} results obtained by varying the spatial stages $S$ on the test set of the MLRetSet \cite{omruuzun2024novel} dataset. $\ac{cr}_\text{spec}$, $\ac{cr}_\text{spat}$ and \ac{cr} denote spectral, spatial and joint spatio-spectral \acl{cr}, respectively. Reconstruction quality is evaluated using \ac{psnr} and \ac{sa}.}
    \label{tab:results-ablation-mlretset}
    \begin{tabular}{S S[round-mode=places,round-precision=2,detect-weight=true] S[round-mode=places,round-precision=2,detect-weight=true] S[round-mode=places,round-precision=2,detect-weight=true] H S[round-mode=places,round-precision=2,detect-weight=true] S[round-mode=places,round-precision=2,detect-weight=true] H H}
        \hline
        $S$ & \ac{cr} & $\ac{cr}_\text{spec}$ & $\ac{cr}_\text{spat}$ & Latent Space & \ac{psnr} $\uparrow$ & \ac{sa} $\downarrow$ & \acs{flops} $\downarrow$ & Parameters $\downarrow$ \\
        \hline
        \SI{0}{\times} & 4.0109 & 4.0109 & 1.0 & \qtyproduct{96x96x92}{} & \textbf{\SI{44.858}{\decibel}} & \textbf{\ang{1.4403}} & \SI{17.44}{\giga\nothing} & \num{946637} \\          
        \SI{1}{\times} & 3.9784 & 0.9946 & 4.0 & \qtyproduct{48x48x371}{} & \SI{42.644}{\decibel} & \ang{1.8489} & \SI{5.66}{\giga\nothing} & \num{1282292} \\
        \SI{2}{\times} & 3.9704 & 0.24815 & 16.0 & \qtyproduct{24x24x1487}{} & \SI{42.279}{\decibel} & \ang{1.8878} & \SI{6.56}{\giga\nothing} & \num{3464848} \\
        \SI{3}{\times} & 3.9704 & 0.0620375 & 64.0 & \qtyproduct{12x12x5951}{} & \SI{40.341}{\decibel} & \ang{2.1518} & \SI{6.78}{\giga\nothing} & \num{7329728} \\
        \hline
        \SI{0}{\times} & 7.8511 & 7.8511 & 1.0 & \qtyproduct{96x96x47}{} & \textbf{\SI{44.893}{\decibel}} & \textbf{\ang{1.4354}} & \SI{15.74}{\giga\nothing} & \num{854432} \\
        \SI{1}{\times} & 7.8095 & 1.952375 & 4.0 & \qtyproduct{48x48x189}{} & \SI{42.617}{\decibel} & \ang{1.8515} & \SI{5.46}{\giga\nothing} & \num{1235518} \\
        \SI{2}{\times} & 7.7787 & 0.48616875 & 16.0 & \qtyproduct{24x24x759}{} & \SI{42.119}{\decibel} & \ang{1.9140} & \SI{6.34}{\giga\nothing} & \num{3277752} \\
        \SI{3}{\times} & 7.7710 & 0.121421875 & 64.0 & \qtyproduct{12x12x3039}{} & \SI{40.975}{\decibel} & \ang{2.0740} & \SI{6.58}{\giga\nothing} & \num{6581344} \\
        \hline
        \SI{0}{\times} & 16.043 & 16.043 & 1.0 & \qtyproduct{96x96x23}{} & \textbf{\SI{44.855}{\decibel}} & \textbf{\ang{1.4418}} & \SI{14.84}{\giga\nothing} & \num{805256} \\
        \SI{1}{\times} & 15.702 & 3.9255 & 4.0 & \qtyproduct{48x48x94}{} & \SI{42.928}{\decibel} & \ang{1.7858} & \SI{5.34}{\giga\nothing} & \num{1211103} \\
        \SI{2}{\times} & 15.578 & 0.973625 & 16.0 & \qtyproduct{24x24x379}{} & \SI{42.234}{\decibel} & \ang{1.9049} & \SI{6.24}{\giga\nothing} & \num{3180092} \\
        \SI{3}{\times} & 15.547 & 0.242921875 & 64.0 & \qtyproduct{12x12x1519}{} & \SI{40.358}{\decibel} & \ang{2.1499} & \SI{6.46}{\giga\nothing} & \num{6190704} \\
        \hline
        \SI{0}{\times} & 30.75 & 30.75 & 1.0 & \qtyproduct{96x96x12}{} & \textbf{\SI{44.786}{\decibel}} & \textbf{\ang{1.4524}} & \SI{14.42}{\giga\nothing} & \num{782.717} \\
        \SI{1}{\times} & 28.941 & 7.23525 & 4.0 & \qtyproduct{48x48x51}{} & \SI{42.625}{\decibel} & \ang{1.8523} & \SI{5.28}{\giga\nothing} & \num{1200052} \\
        \SI{2}{\times} & 28.941 & 1.8088125 & 16.0 & \qtyproduct{24x24x204}{} & \SI{42.270}{\decibel} & \ang{1.8920} & \SI{6.18}{\giga\nothing} & \num{3135117} \\
        \SI{3}{\times} & 28.870 & 0.45109375 & 64.0 & \qtyproduct{12x12x818}{} & \SI{37.633}{\decibel} & \ang{2.7232} & \SI{6.4}{\giga\nothing} & \num{6010547} \\
        \hline
        \SI{0}{\times} & 61.5 & 61.5 & 1.0 & \qtyproduct{96x96x6}{} & \textbf{\SI{44.231}{\decibel}} & \textbf{\ang{1.5482}} & \SI{14.2}{\giga\nothing} & \num{770.423} \\
        \SI{1}{\times} & 61.5 & 15.375 & 4.0 & \qtyproduct{48x48x24}{} & \SI{42.888}{\decibel} & \ang{1.7798} & \SI{5.26}{\giga\nothing} & \num{1193113} \\
        \SI{2}{\times} & 61.5 & 3.84375 & 16.0 & \qtyproduct{24x24x96}{} & \SI{42.279}{\decibel} & \ang{1.8850} & \SI{6.14}{\giga\nothing} & \num{3107361} \\
        \SI{3}{\times} & 61.5 & 0.9609375 & 64.0 & \qtyproduct{12x12x384}{} & \SI{41.282}{\decibel} & \ang{2.0057} & \SI{6.38}{\giga\nothing} & \num{5899009} \\
        \hline
        \SI{0}{\times} & 123.0 & 123.0 & 1.0 & \qtyproduct{96x96x3}{} & \textbf{\SI{42.930}{\decibel}} & \textbf{\ang{1.7953}} & \SI{14.08}{\giga\nothing} & \num{764.276} \\
        \SI{1}{\times} & 123.0 & 30.75 & 4.0 & \qtyproduct{48x48x12}{} & \SI{42.443}{\decibel} & \ang{1.8582} & \SI{5.24}{\giga\nothing} & \num{1190029} \\
        \SI{2}{\times} & 123.0 & 7.6875 & 16.0 & \qtyproduct{24x24x48}{} & \SI{42.269}{\decibel} & \ang{1.8906} & \SI{6.14}{\giga\nothing} & \num{3095025} \\
        \SI{3}{\times} & 123.0 & 1.921875 & 64.0 & \qtyproduct{12x12x192}{} & \SI{41.419}{\decibel} & \ang{1.9732} & \SI{6.36}{\giga\nothing} & \num{5849665} \\
        \hline
        \SI{0}{\times} & 184.5 & 184.5 & 1.0 & \qtyproduct{96x96x2}{} & \SI{40.977}{\decibel} & \ang{2.164} & \SI{14.04}{\giga\nothing} & \num{762227} \\
        \SI{1}{\times} & 184.5 & 46.125 & 4.0 & \qtyproduct{48x48x8}{} & \textbf{\SI{42.403}{\decibel}} & \textbf{\ang{1.8436}} & \SI{5.24}{\giga\nothing} & \num{1189001} \\
        \SI{2}{\times} & 184.5 & 11.53125 & 16.0 & \qtyproduct{24x24x32}{} & \SI{42.169}{\decibel} & \ang{1.9107} & \SI{6.12}{\giga\nothing} & \num{3090913} \\
        \SI{3}{\times} & 184.5 & 2.8828125 & 64.0 & \qtyproduct{12x12x128}{} & \SI{41.126}{\decibel} & \ang{2.0246} & \SI{6.36}{\giga\nothing} & \num{5833217} \\
        \hline
        \SI{0}{\times} & 369.0 & 369.0 & 1.0 & \qtyproduct{96x96x1}{} & \SI{33.548}{\decibel} & \ang{4.8136} & \SI{14.02}{\giga\nothing} & \num{760178} \\
        \SI{1}{\times} & 369.0 & 92.25 & 4.0 & \qtyproduct{48x48x4}{} & \textbf{\SI{42.169}{\decibel}} & \textbf{\ang{1.9021}} & \SI{5.24}{\giga\nothing} & \num{1187973} \\
        \SI{2}{\times} & 369.0 & 23.0625 & 16.0 & \qtyproduct{24x24x16}{} & \SI{42.125}{\decibel} & \ang{1.9105} & \SI{6.12}{\giga\nothing} & \num{3086801} \\
        \SI{3}{\times} & 369.0 & 5.765625 & 64.0 & \qtyproduct{12x12x64}{} & \SI{41.970}{\decibel} & \ang{1.8823} & \SI{6.36}{\giga\nothing} & \num{5816769} \\
        \hline
        \SI{1}{\times} & 738.0 & 184.5 & 4.0 & \qtyproduct{48x48x2}{} & \SI{40.431}{\decibel} & \ang{2.2440} & \SI{5.24}{\giga\nothing} & \num{1187459} \\
        \SI{2}{\times} & 738.0 & 46.125 & 16.0 & \qtyproduct{24x24x8}{} & \SI{41.433}{\decibel} & \ang{2.0202} & \SI{6.12}{\giga\nothing} & \num{3084745} \\
        \SI{3}{\times} & 738.0 & 11.53125 & 64.0 & \qtyproduct{12x12x32}{} & \textbf{\SI{41.574}{\decibel}} & \textbf{\ang{1.9447}} & \SI{6.36}{\giga\nothing} & \num{5808545} \\
        \hline
        \SI{1}{\times} & 1476.0 & 369.0 & 4.0 & \qtyproduct{48x48x1}{} & \SI{37.680}{\decibel} & \ang{2.8319} & \SI{5.22}{\giga\nothing} & \num{1187202} \\
        \SI{2}{\times} & 1476.0 & 92.25 & 16.0 & \qtyproduct{24x24x4}{} & \SI{40.799}{\decibel} & \ang{2.0822} & \SI{6.12}{\giga\nothing} & \num{3083717} \\
        \SI{3}{\times} & 1476.0 & 23.0625 & 64.0 & \qtyproduct{12x12x16}{} & \textbf{\SI{41.037}{\decibel}} & \textbf{\ang{2.0138}} & \SI{6.34}{\giga\nothing} & \num{5804433} \\
        \hline
    \end{tabular}
\end{table}
One can observe that spatio-spectral compression becomes effective primarily at $\acp{cr} > \num{128}$.
This behavior results from the high number of spectral bands (\num{369}) in MLRetSet, which allows spectral compression to maintain its effectiveness across a broader range of \acp{cr} before saturation.
Consequently, the performance disparity between spectral and spatio-spectral compression is less pronounced at lower \acp{cr}.
This suggests that spatial information plays a more significant role in MLRetSet, due to its high spatial resolution.


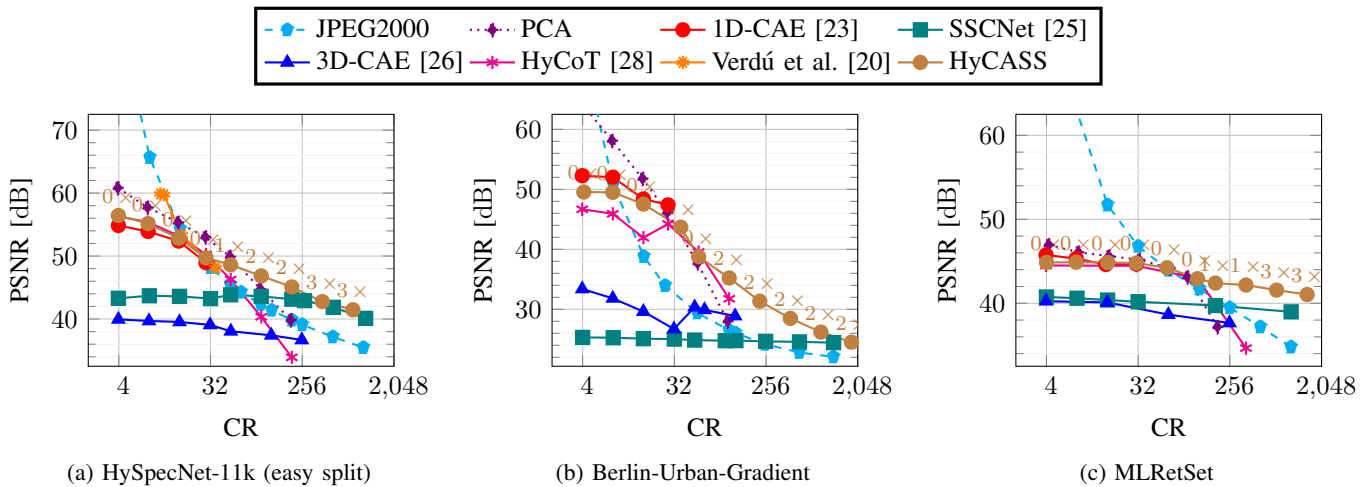
\begin{figure*}
    \centering
    \begin{subfigure}{\textwidth}
        \centering
        \ref{sharedlegend}
        \vspace{1em}
    \end{subfigure}
    \begin{subfigure}{0.32\textwidth}
        \centering
        \begin{tikzpicture}
            \begin{axis}[
                width=\linewidth-4.48941pt,
                height=0.85\linewidth,
                xmode=log,
                log basis x={2},
                log ticks with fixed point,
                minor tick num=4,
                ymin=32.5,
                ymax=72.5,
                xmin=2,
                xmax=2048,
                xticklabel={
                    \pgfkeys{/pgf/fpu=true}
                    \pgfmathparse{int(2^\tick)}
                    \pgfmathprintnumber[fixed]{\pgfmathresult}
                },
                xlabel={CR},
                ylabel={PSNR [\si{\decibel}]},
                legend cell align={left},
                legend pos=north east,
                legend to name=sharedlegend, 
                legend columns=4,
            ]
            \addplot[cyan,dashed,mark=pentagon*] coordinates {
                (4.496926408056514, 84.78510284423828)
                (8.055907758301597, 65.63658905029297)
                (16.002064534761335, 54.256221771240234)
                (28.004665796950693, 49.11704635620117)
                (32.0055978731953, 48.17148208618164)
                (50.00820630746861, 45.49775695800781)
                (64.00993033199593, 44.26495361328125)
                (100.01443666502945, 42.33738327026367)
                (128.01734169296608, 41.40114212036133)
                (202.0133409860396, 39.83381271362305)
                (255.99386549451086, 39.09691619873047)
                (511.89092896530104, 37.16923522949219)
                (1023.2993725386474, 35.469783782958984)
            };
            \addlegendentry{JPEG2000};

            \addplot[violet,dotted,mark=diamond*] coordinates {
                (3.9116120308998594, 60.84214401245117)
                (7.671016461229654, 57.73904037475586)
                (15.334853118339357, 55.34886169433594)
                (28.456183794194526, 53.02031326293945)
                (49.73353770324287, 49.92660903930664)
                (99.16605740995986, 44.76227951049805)
                (197.13890874434122, 39.86271286010742)
            };
            \addlegendentry{\ac{pca}};
            
            \addplot[red,mark=*] coordinates {
                (3.960784314, 54.84735990895165)        
                (7.769230769, 53.902654146940)          
                (15.538461538, 52.38174220872962)       
                (28.857142857, 48.952538482825375)      
            };
            \addlegendentry{\acs{1dcae} \cite{kuester20211d}};

            \addplot[teal,mark=square*] coordinates {
                (3.96078431373, 43.29476347234514)      
                (8, 43.69138220945994)                  
                (15.8431372549, 43.60347966353098)      
                (32, 43.2428080505795)                  
                (50.5, 43.847)
                (101, 43.597)
                (202, 43.149)
                (269.33, 42.918)
                (517.12, 41.855)
                (1077.3, 40.107)
            };
            \addlegendentry{\acs{sscnet} \cite{la2022hyperspectral}};

            \addplot[blue,mark=triangle*] coordinates {
                (3.96078431373, 39.94142961502075)      
                (7.92156862745, 39.6920463376575)       
                (15.8431372549, 39.5384429163403)       
                (31.6862745098, 39.06108974227706)      
                (50.698, 38.052)
                (126.75, 37.414)
                (253.49, 36.675)
            };
            \addlegendentry{\acs{3dcae} \cite{chong2021end}};

            \addplot[magenta,mark=asterisk] coordinates {
                (3.9608, 56.294)                        
                (7.7692, 55.377)                        
                (15.538, 53.202)                        
                (28.857, 50.257)                        
                (50.5, 46.273)                          
                (101, 40.354)                           
                (202, 33.912)                           
            };
            \addlegendentry{\acs{hycot} \cite{fuchs2024hycot}};

            \addplot[orange,mark=10-pointed star] coordinates {
                (10.255691769, 59.86301369863014)
                (11.133079848, 59.726027397260275)
                (16.589235127, 53.35616438356164)
                (35.926380368, 48.150684931506845)
            };
            \addlegendentry{\hl{Verdú et al. {\cite{mijares2023scalable}}}};





            \addplot[brown,mark=*] coordinates {
                (3.9608, 56.444)     
                (7.7692, 55.155)     
                (15.538, 52.828)    
                (28.86, 49.719)    
                (50.50, 48.61)    
                (101.0, 46.84)    
                (202.0, 45.09)    
                (404.0, 42.76)    
                (808.0, 41.46)    
            };
            \addlegendentry{\acs{ours}};

            \node[font=\footnotesize,brown] at (axis cs:3.96, 56.444) [anchor=south] {\SI{0}{\times}};
            \node[font=\footnotesize,brown] at (axis cs:7.77, 55.155) [anchor=south] {\SI{0}{\times}};
            \node[font=\footnotesize,brown] at (axis cs:15.54, 52.828) [anchor=south] {\SI{0}{\times}};
            \node[font=\footnotesize,brown] at (axis cs:28.86, 49.719) [anchor=south] {\SI{0}{\times}};
            \node[font=\footnotesize,brown] at (axis cs:50.50, 48.61) [anchor=south ] {\SI{1}{\times}};
            \node[font=\footnotesize,brown] at (axis cs:101.0, 46.84) [anchor=south] {\SI{2}{\times}};
            \node[font=\footnotesize,brown] at (axis cs:202.0, 45.09) [anchor=south] {\SI{2}{\times}};
            \node[font=\footnotesize,brown] at (axis cs:404.0, 42.76) [anchor=south] {\SI{3}{\times}};
            \node[font=\footnotesize,brown] at (axis cs:808.0, 41.46) [anchor=south] {\SI{3}{\times}};
            \end{axis}
        \end{tikzpicture}
        \caption{HySpecNet-11k (easy split)}
        \label{fig:rate-distortion-plot-hyspecnet11k}
    \end{subfigure}
    \hfill
    \begin{subfigure}{0.32\textwidth}
        \centering
        \begin{tikzpicture}
            \begin{axis}[
                    width=\linewidth-4.48941pt,
                    height=0.85\linewidth,
                    xmode=log,
                    log basis x={2},
                    log ticks with fixed point,
                    minor tick num=4,
                    ymin=20.5,
                    ymax=62.5,
                    xmin=2,
                    xmax=2048,
                    xticklabel={
                        \pgfkeys{/pgf/fpu=true}
                        \pgfmathparse{int(2^\tick)}
                        \pgfmathprintnumber[fixed]{\pgfmathresult}
                    },
                    xlabel={CR},
                    ylabel={PSNR [\si{\decibel}]},
                ]
                \addplot[cyan,dashed,mark=pentagon*] coordinates {
                    (4.000472419602268, 73.64608764648438)
                    (8.00066171781124, 50.77434539794922)
                    (16.00220810435303, 38.79527282714844)
                    (25.999315432981028, 33.954078674316406)
                    (54.004416147426994, 29.397106170654297)
                    (109.95894226953017, 26.532949447631836)
                    (127.96687219116342, 26.046411514282227)
                    (251.6931691630532, 24.166736602783203)
                    (544.681131336308, 22.799592971801758)
                    (1179.38174064588, 22.061893463134766)
                };
                
                \addplot[violet,dotted,mark=diamond*] coordinates {
                    (3.894331182607075, 64.44233703613281)
                    (7.783925929984112, 58.13358688354492)
                    (15.5489406408685, 51.80513000488281)
                    (27.161154654941694, 46.162532806396484)
                    (54.09274347064646, 37.72783279418945)
                    (107.27876774388402, 28.04143524169922)
                };
                
                \addplot[red,mark=*] coordinates {
                    (3.9643, 52.268)
                    (7.9286, 52.015)
                    (15.857, 48.396)
                    (27.750, 47.404)
                };
                
                \addplot[teal,mark=square*] coordinates {
                    (4.0000, 25.306)
                    (8.0000, 25.260)
                    (15.857, 25.102)
                    (32.000, 25.027)
                    (50.743, 24.866)
                    (101.49, 24.764)
                    (129.16, 24.774)
                    (253.71, 24.653)
                    (546.46, 24.559)
                    (1184.0, 24.441)
                };
    
                \addplot[blue,mark=triangle*] coordinates {
                    (3.9643, 33.388)
                    (7.9286, 31.818)
                    (15.857, 29.615)
                    (31.714, 26.709)
                    (50.743, 30.347)
                    (63.429, 29.905)
                    (126.86, 28.871)
                };
    
                \addplot[magenta,mark=asterisk] coordinates {
                    (3.9643, 46.676)
                    (7.9286, 45.920)
                    (15.857, 41.940)
                    (27.750, 44.224)
                    (55.500, 39.545)
                    (111.00, 31.753)
                };
    
                \addplot[brown,mark=*] coordinates {
                    (4.1111, 49.547)
                    (7.9286, 49.499)
                    (15.857, 47.534)
                    (37.000, 43.684)
                    (55.500, 38.768)
                    (111.00, 35.225)
                    (222.00, 31.345)
                    (444.00, 28.484)
                    (888.00, 26.194)
                    (1776.0, 24.519)
                };

                \node[font=\footnotesize,brown] at (axis cs:4.1111, 49.547) [anchor=south] {\SI{0}{\times}};
                \node[font=\footnotesize,brown] at (axis cs:7.9286, 49.499) [anchor=south] {\SI{0}{\times}};
                \node[font=\footnotesize,brown] at (axis cs:15.587, 47.534) [anchor=south] {\SI{0}{\times}};
                \node[font=\footnotesize,brown] at (axis cs:37.000, 43.684) [anchor=south] {\SI{0}{\times}};
                \node[font=\footnotesize,brown] at (axis cs:55.500, 38.768) [anchor=south] {\SI{0}{\times}};
                \node[font=\footnotesize,brown] at (axis cs:111.00, 35.225) [anchor=south] {\SI{2}{\times}};
                \node[font=\footnotesize,brown] at (axis cs:222.00, 31.345) [anchor=south] {\SI{2}{\times}};
                \node[font=\footnotesize,brown] at (axis cs:444.00, 28.484) [anchor=south] {\SI{2}{\times}};
                \node[font=\footnotesize,brown] at (axis cs:888.00, 26.194) [anchor=south] {\SI{2}{\times}};
                \node[font=\footnotesize,brown] at (axis cs:1776.0, 24.519) [anchor=south] {\SI{2}{\times}};
            \end{axis}
        \end{tikzpicture}
        \caption{\hl{Berlin-Urban-Gradient}}
        \label{fig:rate-distortion-plot-bug}
    \end{subfigure}
    \hfill
    \begin{subfigure}{0.32\textwidth}
        \centering
        \begin{tikzpicture}
            \begin{axis}[
                    width=\linewidth-4.48941pt,
                    height=0.85\linewidth,
                    xmode=log,
                    log basis x={2},
                    log ticks with fixed point,
                    minor tick num=4,
                    ymin=32.5,
                    ymax=62.5,
                    xmin=2,
                    xmax=2048,
                    xticklabel={
                        \pgfkeys{/pgf/fpu=true}
                        \pgfmathparse{int(2^\tick)}
                        \pgfmathprintnumber[fixed]{\pgfmathresult}
                    },
                    xlabel={CR},
                    ylabel={PSNR [\si{\decibel}]},
                ]
                \addplot[cyan,dashed,mark=pentagon*] coordinates {
                    (4.0026843771946075, 80.45751953125)
                    (8.000290205879358, 62.84672546386719)
                    (16.00153130503918, 51.70823287963867)
                    (32.00380074072066, 46.77833557128906)
                    (64.00925486384209, 43.94217300415039)
                    (128.00834985490735, 41.71170425415039)
                    (255.96729842104898, 39.466590881347656)
                    (511.8083462658521, 37.18948745727539)
                    (1023.1360914819182, 34.7945671081543)
                };
                
                \addplot[violet,dotted,mark=diamond*] coordinates {
                    (4.201238956027217, 46.9564323425293)
                    (8.223734659241973, 46.0838737487793)
                    (16.43240237332737, 45.62356185913086)
                    (32.80470229068165, 45.18788146972656)
                    (65.37030891996525, 44.1751594543457)
                    (97.69942435424355, 43.125091552734375)
                    (193.29330685667563, 37.20418167114258)
                };
                
                \addplot[red,mark=*] coordinates {
                    (3.9677, 45.757)        
                    (7.8511, 45.318)        
                    (15.375, 44.620)        
                    (30.750, 44.671)        
                };
                
                \addplot[teal,mark=square*] coordinates {
                    (4.0, 40.749)
                    (8.0, 40.603)
                    (16.0, 40.384)
                    (32.0, 40.182)
                    (184.5, 39.729)
                    (1026.8, 38.979)
                };
    
                \addplot[blue,mark=triangle*] coordinates {
                    (3.9677, 40.245)
                    (15.871, 40.102)
                    (63.484, 38.651)
                    (253.94, 37.654)
                };
    
                \addplot[magenta,mark=asterisk] coordinates {
                    (3.9677, 44.502)
                    (30.750, 44.464)
                    (123.0, 42.975)
                    (369.0, 34.657)
                };
    
    
                \addplot[brown,mark=*] coordinates {
                    (4.01, 44.86)
                    (7.85, 44.89)
                    (16.04, 44.86)
                    (30.75, 44.79)
                    (61.5, 44.23)
                    (123, 42.93)
                    (184.5, 42.40)
                    (369, 42.17)
                    (738, 41.57)
                    (1476, 41.04)
                };
    
                \node[font=\footnotesize,brown] at (axis cs:4.01, 44.86) [anchor=south] {\SI{0}{\times}};
                \node[font=\footnotesize,brown] at (axis cs:7.85, 44.89) [anchor=south] {\SI{0}{\times}};
                \node[font=\footnotesize,brown] at (axis cs:16.04, 44.86) [anchor=south] {\SI{0}{\times}};
                \node[font=\footnotesize,brown] at (axis cs:30.75, 44.79) [anchor=south] {\SI{0}{\times}};
                \node[font=\footnotesize,brown] at (axis cs:61.5, 44.23) [anchor=south] {\SI{0}{\times}};
                \node[font=\footnotesize,brown] at (axis cs:123, 42.93) [anchor=south] {\SI{0}{\times}};
                \node[font=\footnotesize,brown] at (axis cs:184.5, 42.40) [anchor=south] {\SI{1}{\times}};
                \node[font=\footnotesize,brown] at (axis cs:369, 42.17) [anchor=south] {\SI{1}{\times}};
                \node[font=\footnotesize,brown] at (axis cs:738, 41.57) [anchor=south] {\SI{3}{\times}};
                \node[font=\footnotesize,brown] at (axis cs:1476, 41.04) [anchor=south] {\SI{3}{\times}};
            
            \end{axis}
        \end{tikzpicture}
        \caption{MLRetSet}
        \label{fig:rate-distortion-plot-mlretset}
    \end{subfigure}
    \caption{Rate-distortion performance on the test set of (\subref{fig:rate-distortion-plot-hyspecnet11k}) HySpecNet-11k \cite{fuchs2023hyspecnet} (easy split)\hl{, {(\subref{fig:rate-distortion-plot-bug})} Berlin-Urban-Gradient {\cite{okujeni2016berlin}}} and (\subref{fig:rate-distortion-plot-mlretset}) MLRetSet \cite{omruuzun2024novel}. Rate is visualized as \ac{cr} and distortion is given as \ac{psnr} in \ac{decibel}.}
    \label{fig:rate-distortion-plot}
\end{figure*}

\subsection{Comparison with Other Approaches}
\label{subsec:experimental-results-sota}
This subsection analyzes the effectiveness of \ac{ours} in terms of \ac{psnr} at different \acp{cr} comparing it with several traditional baselines and state-of-the-art learning-based \ac{hsi} compression models on the HySpecNet-11k\hl{, the Berlin-Urban-Gradient} and the MLRetSet datasets.
The comparative models include:
\begin{enumerate*}[1)]
    \item JPEG2000;
    \item \ac{pca};
    \item \ac{1dcae} \cite{kuester20211d};
    \item \ac{sscnet} \cite{la2022hyperspectral};
    \item \ac{3dcae} \cite{chong2021end};
    \item \ac{hycot} \cite{fuchs2024hycot}; and
    \item \hl{Verdú et al. {\cite{mijares2023scalable}}}.
\end{enumerate*}
\autoref{fig:rate-distortion-plot} illustrates the corresponding rate-distortion curves, where the rate is expressed as the \ac{cr} and distortion is measured as \ac{psnr} in \ac{decibel}.

\subsubsection{HySpecNet-11k}
\autoref{fig:rate-distortion-plot} (\subref{fig:rate-distortion-plot-hyspecnet11k}) shows that our proposed model achieves superior \ac{psnr} reconstruction quality on the HySpecNet-11k dataset (low spatial resolution) across nearly all \acp{cr} when compared to state-of-the-art learning-based models.
\hl{Only the method from Verdú et al. {\cite{mijares2023scalable}} achieves a higher {\ac{psnr}} for {\acp{cr}} between {\num{8}} and {\num{16}}, despite focusing on the compression of spatial redundancies, due to its particular architectural design for this compression range.}
Moreover, traditional methods demonstrate superior performance at $\acp{cr} < \num{64}$.
JPEG2000 performs better for \acp{cr} below \num{16}, while \ac{pca} gives higher \ac{psnr} value even for \acp{cr} up to \num{64}.
\hl{These results suggest that for ${\acp{cr}} < {\num{64}}$ traditional methods remain more effective, and the increased complexity of learning-based models may offer limited benefits in this {\ac{cr}} range. Furthermore it shows that spectral compression is preferable to spatial compression for low {\acp{cr}} when the spatial resolution is low.}
At a $\ac{cr} \approx \num{101}$, \ac{ours} with two spatial stages \hl{(spatio-spectral compression)} reaches a \ac{psnr} of \SI{46.84}{\decibel}, clearly surpassing the best-performing state-of-the-art model \ac{sscnet} \hl{(spatial compression)}, which achieves \SI{43.597}{\decibel} and also the \ac{pca} baseline \hl{(spectral compression)}, which achieves \SI{44.76}{\decibel}.
\hl{This suggests that spatio-spectral compression is beneficial when dealing with medium {\acp{cr}} and low spatial resolution.}
Similarly, at the \ac{cr} of approximately \num{1024}, \ac{ours} achieves a \ac{psnr} of \SI{41.46}{\decibel}, outperforming \ac{sscnet} and JPEG2000 that reach \SI{40.11}{\decibel} and \SI{35.47}{\decibel}, respectively.
For $\acp{cr} > \num{64}$, our results demonstrate the superior effectiveness of our proposed  model over the other learning-based models and traditional approaches\hl{, due to {\ac{ours}}'s adjustable spatio-spectral compression}.
A comparative analysis of state-of-the-art learning-based models reveals that for $\acp{cr} < \num{32}$, our proposed model, which exclusively performs spectral compression in this range without any spatial stages, shows limited improvement over the state of the art.
This is because \ac{ours} is not specifically optimized for this \ac{cr} range, {\hl{where solely spectral compression is sufficient,}} resulting in performance closely similar to \ac{hycot}, {\hl{a spectral compression model}}.
For $\acp{cr} > \num{256}$, \ac{ours}'s performance converges with that of \ac{sscnet}, a spatial compression model.
\ac{sscnet} naturally excels under these conditions due to its design for strong spatial compression.
It is worth noting that the performance of \ac{ours} at $\acp{cr} > \num{256}$ could potentially be further enhanced by integration of additional spatial stages (e.g., beyond three), which would allow for an even more effective exploitation of spatial redundancies.

\subsubsection{\hl{Berlin-Urban-Gradient}}
\hl{\mbox{\autoref{fig:rate-distortion-plot}} (\mbox{\subref{fig:rate-distortion-plot-bug}}) presents the experimental results on the Berlin-Urban-Gradient dataset, which has a medium spatial resolution of {\SI{3.6}{\meter}}.
For {\acp{cr}} below {\num{16}}, {\ac{pca}} demonstrates superior performance, while JPEG2000 remains competitive for {\acp{cr}} below {\num{8}}.
These observations suggest that, at low {\acp{cr}}, traditional methods are more effective than learning-based models, potentially due to the limited size of the dataset, which makes the training of learning-based models difficult.
{\ac{sscnet}} performs poorly across all tested {\acp{cr}}, and the {\ac{3dcae}} shows limited improvement compared to that, saturating at relatively low ${\acp{psnr}} < {\SI{35}{\decibel}}$.
This highlights the need for a careful design when incorporating spatial redundancies.
In contrast, the {\ac{1dcae}}, which captures short-range spectral dependencies, outperforms {\ac{hycot}}, whose reliance on long-range dependencies is less effective in this case due to the comparatively low number of only {\SI{111}{\sband}}.
Finally, {\ac{ours}} achieves clear benefits for ${\acp{cr}} > {\num{64}}$ by leveraging two spatial stages to achieve spatio-spectral compression.}

\subsubsection{MLRetSet}
\autoref{fig:rate-distortion-plot} (\subref{fig:rate-distortion-plot-mlretset}) presents the rate-distortion curves on the MLRetSet dataset, from which the following observations can be made:
Traditional approaches are highly effective in low-compression regimes.
At $\acp{cr} < \num{64}$, JPEG2000 performs particularly well, surpassing all learning-based models and the \ac{pca} baseline.
The reconstruction quality achieved by the spectral learning-based models \ac{1dcae}, \ac{hycot}, and \ac{ours} with zero spatial stages shows only minor variation, indicating comparable performance for long-range and short-range spectral redundancy compression.
\ac{sscnet} and \ac{3dcae}, which incorporate spatial compression, perform worse than the spectral compression models, especially for $\acp{cr} < \num{64}$.
This suggests that, when sufficient bitrate is available, exploiting spectral redundancies is more straightforward and yields better compression performance than incorporating spatial information.
\ac{ours} achieves comparable performance to learning-based spectral compression models at $\acp{cr} < \num{128}$, while at $\acp{cr} > \num{128}$ it demonstrates advantages thanks to its adjustable design by integrating spatial compression.


\subsection{Visual Analysis}
\begin{figure*}
    \centering
    \begin{tabular}{c|ccccc}
        \raisebox{-.1\height}{\includegraphics[width=0.118\textwidth]{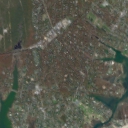}} & \multicolumn{5}{c}{
            \begin{tikzpicture}
                \begin{axis}[
                    hide axis,
                    scale only axis,
                    width=0.65\textwidth,
                    height=10mm,
                    ymin=0, ymax=1,
                    colormap={jet}{
                        rgb=(0,0,0.5)
                        rgb=(0,0,1)
                        rgb=(0,1,1)
                        rgb=(1,1,0)
                        rgb=(1,0,0)
                        rgb=(0.5,0,0)
                    },
                    colorbar,
                    point meta min=0,
                    point meta max=10,
                    colorbar horizontal,
                    colorbar style={
                        width=0.65\textwidth,
                        height=5mm,
                        xtick={0,2,...,10},
                        xticklabel=\pgfmathprintnumber{\tick}$^\circ$,
                        title=\hl{Pixelwise} \ac{sa},
                        title style={
                            at={(0.5,-1.2)},
                            anchor=north,
                        },
                    },
                ]
                    \addplot[draw=none] coordinates {(0,0) (10,0)};
                \end{axis}
            \end{tikzpicture}
        } \\

        \cline{2-6}
        
        Original \ac{hsi} & \acs{1dcae} \cite{kuester20211d} & \acs{sscnet} \cite{la2022hyperspectral} & \acs{3dcae} \cite{chong2021end} & \acs{hycot} \cite{fuchs2024hycot} & \ac{ours} \\
        \hline \rule{0pt}{7.5ex}

        \multirow{1}{*}{}
            $\ac{cr} \approx \num{4}$ & \raisebox{-.5\height}{\includegraphics[width=0.118\textwidth]{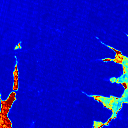}}
            & \raisebox{-.5\height}{\includegraphics[width=0.118\textwidth]{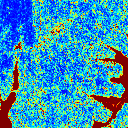}}
            & \raisebox{-.5\height}{\includegraphics[width=0.118\textwidth]{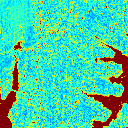}}
            & \raisebox{-.5\height}{\includegraphics[width=0.118\textwidth]{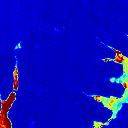}}
            & \raisebox{-.5\height}{\includegraphics[width=0.118\textwidth]{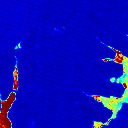}} \\
        \footnotesize \hl{{\acs{sa}}$\downarrow$/{\ac{psnr}}$\uparrow$/{\acs{ssim}}$\uparrow$} & \footnotesize \hl{{\ang{1.18}}/{\SI{55.19}{\decibel}}/\textbf{{\num{1.00}}}} & \footnotesize \hl{{\ang{5.00}}/{\SI{37.53}{\decibel}}/{\num{0.92}}} & \footnotesize \hl{{\ang{5.92}}/{\SI{36.97}{\decibel}}/{\num{0.93}}} & \footnotesize \hl{{\ang{1.12}}/{\SI{56.30}{\decibel}}/\textbf{{\num{1.00}}}} & \footnotesize \hl{\textbf{{\ang{1.11}}}/\textbf{{\SI{56.50}{\decibel}}}/\textbf{{\num{1.00}}}} \\
        \hline \rule{0pt}{7.5ex}
        
        \multirow{1}{*}{}
            $\ac{cr} \approx \num{32}$ & \raisebox{-.5\height}{\includegraphics[width=0.118\textwidth]{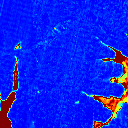}}
            & \raisebox{-.5\height}{\includegraphics[width=0.118\textwidth]{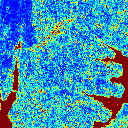}}
            & \raisebox{-.5\height}{\includegraphics[width=0.118\textwidth]{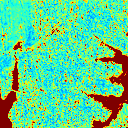}}
            & \raisebox{-.5\height}{\includegraphics[width=0.118\textwidth]{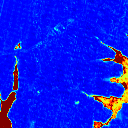}}
            & \raisebox{-.5\height}{\includegraphics[width=0.118\textwidth]{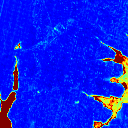}} \\
        \footnotesize \hl{{\acs{sa}}$\downarrow$/{\ac{psnr}}$\uparrow$/{\acs{ssim}}$\uparrow$} & \footnotesize \hl{{\ang{2.35}}/{\SI{48.04}{\decibel}}/{\num{0.99}}} & \footnotesize \hl{{\ang{5.04}}/{\SI{37.64}{\decibel}}/{\num{0.92}}} & \footnotesize \hl{{\ang{6.47}}/{\SI{31.69}{\decibel}}/{\num{0.90}}} & \footnotesize \hl{\textbf{{\ang{2.13}}}/\textbf{{\SI{49.11}{\decibel}}}/\textbf{{\num{1.00}}}} & \footnotesize \hl{{\ang{2.15}}/{\SI{48.81}{\decibel}}/{\num{0.99}}} \\
        \hline \rule{0pt}{7.5ex}
        
        \multirow{1}{*}{}
            $\ac{cr} \approx \num{101}$ &
            & \raisebox{-.5\height}{\includegraphics[width=0.118\textwidth]{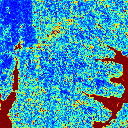}}
            & \raisebox{-.5\height}{\includegraphics[width=0.118\textwidth]{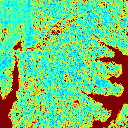}}
            & \raisebox{-.5\height}{\includegraphics[width=0.118\textwidth]{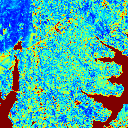}}
            & \raisebox{-.5\height}{\includegraphics[width=0.118\textwidth]{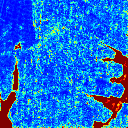}} \\
        \footnotesize \hl{{\acs{sa}}$\downarrow$/{\ac{psnr}}$\uparrow$/{\acs{ssim}}$\uparrow$} & & \footnotesize \hl{{\ang{5.12}}/{\SI{37.39}{\decibel}}/{\num{0.92}}} & \footnotesize \hl{{\ang{7.26}}/{\SI{33.95}{\decibel}}/{\num{0.85}}} & \footnotesize \hl{{\ang{6.47}}/{\SI{38.70}{\decibel}}/{\num{0.96}}} & \footnotesize \hl{\textbf{{\ang{3.70}}}/\textbf{{\SI{42.49}{\decibel}}}/\textbf{{\num{0.98}}}} \\
        \hline \rule{0pt}{7.5ex}
        
        \multirow{1}{*}{}
            $\ac{cr} \approx \num{404}$ &
            & \raisebox{-.5\height}{\includegraphics[width=0.118\textwidth]{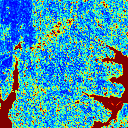}}
            & 
            & 
            & \raisebox{-.5\height}{\includegraphics[width=0.118\textwidth]{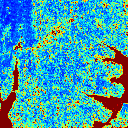}} \\
        \footnotesize \hl{{\acs{sa}}$\downarrow$/{\ac{psnr}}$\uparrow$/{\acs{ssim}}$\uparrow$} & & \footnotesize \hl{{\ang{5.35}}/{\SI{36.79}{\decibel}}/{\num{0.91}}} & & & \footnotesize \hl{\textbf{{\ang{5.20}}}/\textbf{{\SI{36.98}{\decibel}}}/\textbf{{\num{0.92}}}} \\

    \end{tabular}
    \caption{\Ac{sa} error maps of state-of-the-art learning-based \ac{hsi} compression models across three \acp{cr}, evaluated on an example image from the HySpecNet-11k \cite{fuchs2023hyspecnet} dataset.}
    \label{fig:error-maps}
\end{figure*}
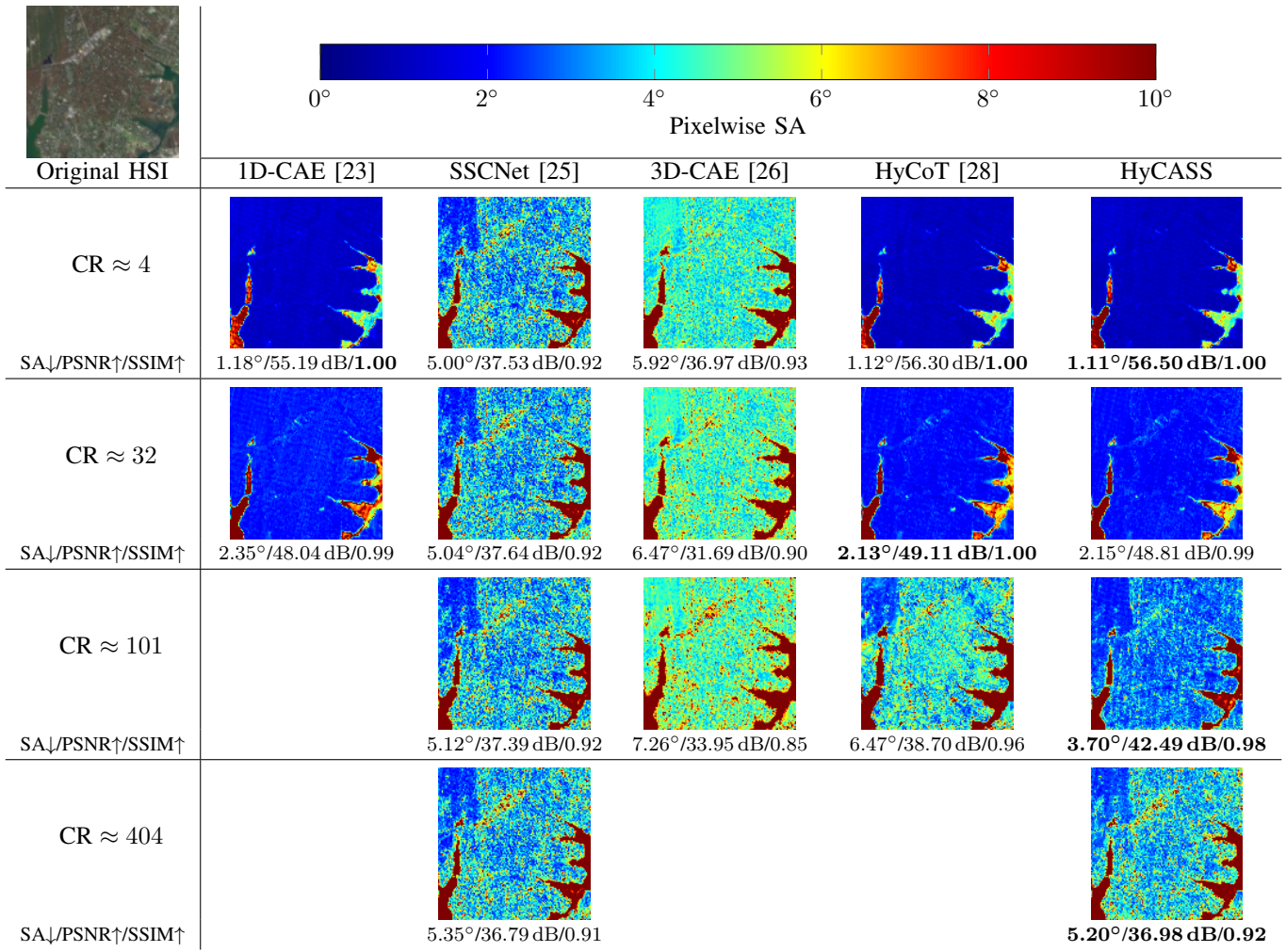

\begin{figure*}
    \centering
    \begin{tabular}{c|ccccc}
        \raisebox{-.0\height}{\includegraphics[width=.093\textwidth]{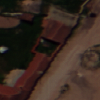}} & \multicolumn{5}{c}{
            \begin{tikzpicture}
                \begin{axis}[
                    hide axis,
                    scale only axis,
                    width=0.65\textwidth,
                    height=10mm,
                    ymin=0, ymax=1,
                    colormap={jet}{
                        rgb=(0,0,0.5)
                        rgb=(0,0,1)
                        rgb=(0,1,1)
                        rgb=(1,1,0)
                        rgb=(1,0,0)
                        rgb=(0.5,0,0)
                    },
                    colorbar,
                    point meta min=0,
                    point meta max=10,
                    colorbar horizontal,
                    colorbar style={
                        width=0.65\textwidth,
                        height=5mm,
                        xtick={0,2,...,10},
                        xticklabel=\pgfmathprintnumber{\tick}$^\circ$,
                        title=Pixelwise \ac{sa},
                        title style={
                            at={(0.5,-1.2)},
                            anchor=north,
                        },
                    },
                ]
                    \addplot[draw=none] coordinates {(0,0) (10,0)};
                \end{axis}
            \end{tikzpicture}
        } \\

        \cline{2-6}
        
        Original \ac{hsi} & \acs{1dcae} \cite{kuester20211d} & \acs{sscnet} \cite{la2022hyperspectral} & \acs{3dcae} \cite{chong2021end} & \acs{hycot} \cite{fuchs2024hycot} & \ac{ours} \\
        \hline \rule{0pt}{7.5ex}

        \multirow{1}{*}{}
            $\ac{cr} \approx \num{4}$ & \raisebox{-.5\height}{\includegraphics[width=0.093\textwidth]{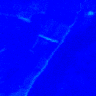}}
            & \raisebox{-.5\height}{\includegraphics[width=0.093\textwidth]{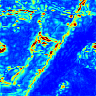}}
            & \raisebox{-.5\height}{\includegraphics[width=0.093\textwidth]{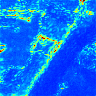}}
            & \raisebox{-.5\height}{\includegraphics[width=0.093\textwidth]{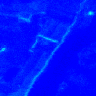}}
            & \raisebox{-.5\height}{\includegraphics[width=0.093\textwidth]{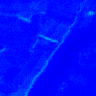}} \\
        \footnotesize \hl{{\acs{sa}}$\downarrow$/{\ac{psnr}}$\uparrow$/{\acs{ssim}}$\uparrow$} & \footnotesize \hl{\textbf{{\ang{1.21}}}/\textbf{{\SI{44.99}{\decibel}}}/\textbf{{\num{0.99}}}} & \footnotesize \hl{{\ang{2.76}}/{\SI{35.01}{\decibel}}/{\num{0.91}}} & \footnotesize \hl{{\ang{2.51}}/{\SI{36.92}{\decibel}}/{\num{0.95}}} & \footnotesize \hl{{\ang{1.43}}/{\SI{43.62}{\decibel}}/{\num{0.98}}} & \footnotesize \hl{{\ang{1.37}}/{\SI{43.95}{\decibel}}/{\num{0.98}}} \\
        \hline \rule{0pt}{7.5ex}
        
        \multirow{1}{*}{}
            $\ac{cr} \approx \num{256}$ & 
            & \raisebox{-.5\height}{\includegraphics[width=0.093\textwidth]{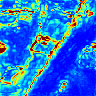}}
            & \raisebox{-.5\height}{\includegraphics[width=0.093\textwidth]{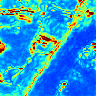}}
            & \raisebox{-.5\height}{\includegraphics[width=0.093\textwidth]{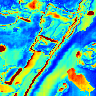}}
            & \raisebox{-.5\height}{\includegraphics[width=0.093\textwidth]{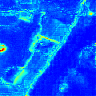}} \\
        \footnotesize \hl{{\acs{sa}}$\downarrow$/{\ac{psnr}}$\uparrow$/{\acs{ssim}}$\uparrow$} & & \footnotesize \hl{{\ang{3.12}}/{\SI{33.42}{\decibel}}/{\num{0.88}}} & \footnotesize \hl{{\ang{3.49}}/{\SI{32.62}{\decibel}}/{\num{0.88}}} & \footnotesize \hl{{\ang{4.34}}/{\SI{32.53}{\decibel}}/{\num{0.91}}} & \footnotesize \hl{\textbf{{\ang{2.17}}}/\textbf{{\SI{39.45}{\decibel}}}/\textbf{{\num{0.97}}}} \\
        \hline \rule{0pt}{7.5ex}
        \multirow{1}{*}{}
            $\ac{cr} \approx \num{1024}$ &
            & \raisebox{-.5\height}{\includegraphics[width=0.093\textwidth]{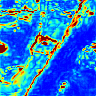}}
            & 
            & 
            & \raisebox{-.5\height}{\includegraphics[width=0.093\textwidth]{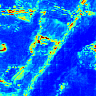}} \\
        \footnotesize \hl{{\acs{sa}}$\downarrow$/{\ac{psnr}}$\uparrow$/{\acs{ssim}}$\uparrow$} & & \footnotesize \hl{{\ang{3.27}}/{\SI{32.52}{\decibel}}/{\num{0.86}}} & & & \footnotesize \hl{\textbf{{\ang{2.56}}}/\textbf{{\SI{36.76}{\decibel}}}/\textbf{{\num{0.94}}}} \\
    \end{tabular}
    \caption{\hl{{\Ac{sa}} error maps of state-of-the-art learning-based {\ac{hsi}} compression models across three {\acp{cr}}, evaluated on an example image from the MLRetSet {\cite{omruuzun2024novel}} dataset.}}
    \label{fig:error-maps-mlretset}
\end{figure*}

For a qualitative evaluation, the reconstruction outputs of the considered learning-based compression models are visually compared \hl{under varying {\acp{cr}} on two exemplary {\acp{hsi}}, which include different land cover types, in {\autoref{fig:error-maps}} and {\autoref{fig:error-maps-mlretset}}.}
The error maps show the reconstruction error (derived from the \ac{sa} for each pixel) across several representative \acp{cr}.
This provides a detailed visual assessment of the spatial distribution of reconstruction errors.
Each case also reports the corresponding \ac{cr}, the overall \hl{{\ac{sa}},} \ac{psnr} \hl{and {\ac{ssim}}} for comprehensive comparison.

\autoref{fig:error-maps} presents the error maps for a reconstructed HySpecNet-11k image\hl{, covering mostly urban, water, and forest areas}.
At $\ac{cr} \approx \num{4}$, the spectral compression models \ac{1dcae}, \ac{hycot}, and \ac{ours} with zero spatial stages achieve relatively low \ac{sa} errors \hl{of slightly above {\ang{1}}} across most of the scene, as also indicated by their high overall \ac{psnr} values exceeding \SI{55}{\decibel} \hl{and a {\ac{ssim}} of {\num{1.00}}}.
Urban areas are reconstructed with high precision, showing $\ac{sa} < \ang{2}$, while water regions pose more difficulty, with errors exceeding $\ac{sa} > \ang{5}$.
In contrast, \ac{sscnet} and \ac{3dcae} achieve lower \ac{psnr} values of approximately \SI{37}{\decibel} \hl{and also the {\ac{ssim}} is decreasing to {\num{0.92}} and {\num{0.93}}, respectively}.
Consequently, their error maps contain considerable noise in the urban regions.
Similar to spectral compression models, these models also face difficulties in reconstructing water areas, whereas the forest region in the top-left shows the highest reconstruction fidelity.
This suggests that at such low \ac{cr}, spatial compression adversely affects reconstruction quality, limiting the model's ability to preserve fine spectral details of each pixel due to the spatial downsampling.
At $\ac{cr} \approx \num{32}$, distortion increases for most methods.
\ac{sscnet} shows a slight improvement in reconstruction quality compared to its performance at $\ac{cr} \approx \num{4}$, although overall quality remains low.
Additionally, the models \ac{1dcae}, \ac{hycot}, and \ac{ours} exhibit increased distortion in urban regions, with notably elevated \ac{sa} values observed in the industrial area situated at the top center of the \ac{hsi}.
At $\ac{cr} \approx \num{101}$, reconstruction degradation becomes more visible.
While \ac{sscnet} and \ac{3dcae} exhibit minimal differences to lower \acp{cr}, the spectral compression model \ac{hycot} exhibits substantial errors in urban and water regions, however it maintains relatively strong performance in forested areas.
In contrast, \ac{ours}, employing two spatial stages at this \ac{cr}, exhibits increased error in urban regions but it achieves the highest overall \hl{{\ac{sa}}, {\ac{psnr}} and {\ac{ssim}}}.
\hl{Further degradation can be observed at ${\ac{cr}} \approx \num{404}$.}

\hl{{\autoref{fig:error-maps-mlretset}} shows the error maps of a reconstructed MLRetSet image, which includes mainly buildings, sand roads, and grassland areas,  for three different {\acp{cr}}.}
\hl{At ${\ac{cr}} \approx {\num{4}}$, the spectral compression models {\ac{1dcae}}, {\ac{hycot}}, and {\ac{ours}} with zero spatial stages exhibit reconstruction errors primarily in the shadow regions of buildings while urban areas and grassland are reconstructed with high accuracy (${\ac{sa}} < {\ang{2}}$).
{\ac{sscnet}} and {\ac{3dcae}} exhibit their major reconstruction errors along the object boundaries of buildings due to their spatial compression.
At ${\ac{cr}} \approx {\num{256}}$, {\ac{ours}} with a single spatial stage demonstrates an increase in {\ac{sa}} for urban regions, while sand roads and grassland areas remain comparatively stable.
In contrast, {\ac{sscnet}} and {\ac{3dcae}} display localized degradation that extends further into the buildings while {\ac{hycot}} yields a more homogeneous error distribution across the entire {\ac{hsi}}, which shows the limit of spectral compression models when reaching higher {\acp{cr}}.
At ${\ac{cr}} \approx {\num{1024}}$, reconstruction quality continues to decrease in the building areas while grassland and the sand road areas retain a low reconstruction error. 
}

The qualitative results indicate that learning-based models struggle to reconstruct water, shadow and urban regions.
Urban areas become more challenging to reconstruct at higher \acp{cr} and when spatial compression is applied.
In contrast, forestry regions and vegetated areas are consistently reconstructed with the highest quality, due to their frequent representation and homogeneous spatial and spectral characteristics in the training data, which facilitate accurate learning and reconstruction.
Overall, \ac{ours} consistently achieves an effective trade-off between \ac{cr} and reconstruction fidelity across a wide range of \acp{cr} and land cover types.

\section{Conclusion and Discussion}
\label{sec:conclusion}
In this paper, we have introduced \ac{ours}, a novel learning-based \ac{hsi} compression model designed for adjustable spatio-spectral \ac{hsi} compression.
To this end, the proposed model employs six modules:
\begin{enumerate*}[i)]
    \item a spectral encoder module;
    \item a spatial encoder module;
    \item a \ac{cr} adapter encoder module;
    \item a \ac{cr} adapter decoder module;
    \item a spatial decoder module; and
    \item a spectral decoder module.
\end{enumerate*}
Our model accomplishes:
\begin{enumerate*}[1)]
    \item spectral feature extraction (realized within the spectral encoder and decoder modules);
    \item spatial compression with variable stages (realized within the spatial encoder  and decoder modules); and
    \item spectral compression with variable output channels (realized within the \ac{cr} adapter encoder and \ac{cr} decoder modules).
\end{enumerate*}
Unlike existing learning-based \ac{hsi} compression models, \ac{ours} provides flexible control over the trade-off between spectral and spatial compression through its modular design.
We have conducted extensive experiments on \hl{three} benchmark datasets, including ablation studies, comparisons with state-of-the-art methods, and visual analyses of reconstruction errors.
Our results demonstrate the effectiveness of \ac{ours}, and reveal how the balance between spectral and spatial compression affects reconstruction fidelity across different \acp{cr} \hl{and spatial resolutions}.
Our findings confirm the importance of adjustable spatio-spectral compression in addressing the diverse characteristics of hyperspectral data.
It is worth emphasizing that with the continuous growth of hyperspectral data archives, spatio–spectral learning-based \acp{hsi} compression is becoming increasingly important, as it enables significantly higher \acp{cr}. In this context, the proposed model offers a promising solution for efficient and flexible \ac{hsi} compression.
Based on our analyses, we have also derived a guideline to select the trade-off between spectral or spatial compression depending on the spatial resolution and overall \ac{cr} as follows:
\begin{itemize}
    \item For \ac{hsi} data with low spatial resolution: use spectral compression at low \acp{cr}; spatio-spectral compression with greater spectral emphasis at medium \acp{cr}; and spatio-spectral compression with greater spatial emphasis at high \acp{cr}.
    \item \hl{For {\ac{hsi}} data with medium spatial resolution: use spectral compression at low {\acp{cr}}; and balanced spatio-spectral compression at medium and high {\acp{cr}}.}
    \item For \ac{hsi} data with high spatial resolution: use spectral compression at low \acp{cr}; and spatio-spectral compression with greater spatial emphasis at medium and high \acp{cr}.

\end{itemize}

As a final remark, we would like to note that the development of \acp{fm} has attracted great attention in \ac{rs}.
\acp{fm} are usually pre-trained on large-scale datasets and then fine-tuned for specific downstream tasks such as image classification, segmentation, or change detection.
We believe that leveraging the generalization capabilities of \acp{fm} for \ac{hsi} compression can lead to a more robust and effective compression performance.
As a future work, we plan to investigate the use of \acp{fm} as a backbone for \ac{hsi} compression to improve reconstruction fidelity and enable adaptation across diverse sensor and acquisition conditions.


\bibliographystyle{IEEEtran}
\bibliography{bib/refs.bib}

\begin{IEEEbiography}[{\includegraphics[width=1in,height=1.25in,clip,keepaspectratio]{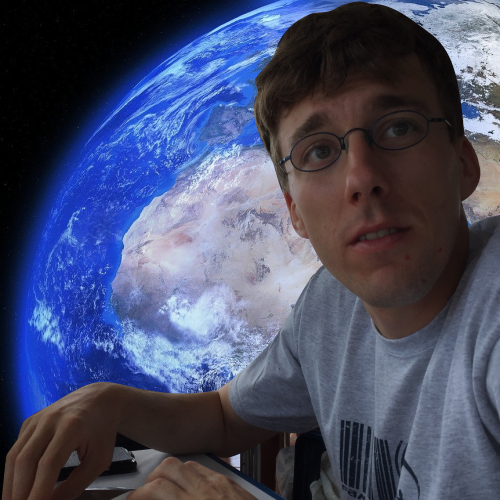}}]{Martin Hermann Paul Fuchs}
received his B. Sc. and M. Sc. degrees in electrical engineering from Technische Universität Berlin (TU Berlin), Berlin, Germany in 2018 and 2021, respectively. He is currently pursuing a Ph. D. degree with the Remote Sensing Image Analysis (RSiM) group at the Faculty of Electrical Engineering and Computer Science, TU Berlin and the Berlin Institute for the Foundations of Learning and Data (BIFOLD). His research interests revolve around the intersection of remote sensing and deep learning, and he has a particular interest in hyperspectral imaging and compression.
\end{IEEEbiography}

\begin{IEEEbiography}[{\includegraphics[width=1in,height=1.25in,clip,keepaspectratio]{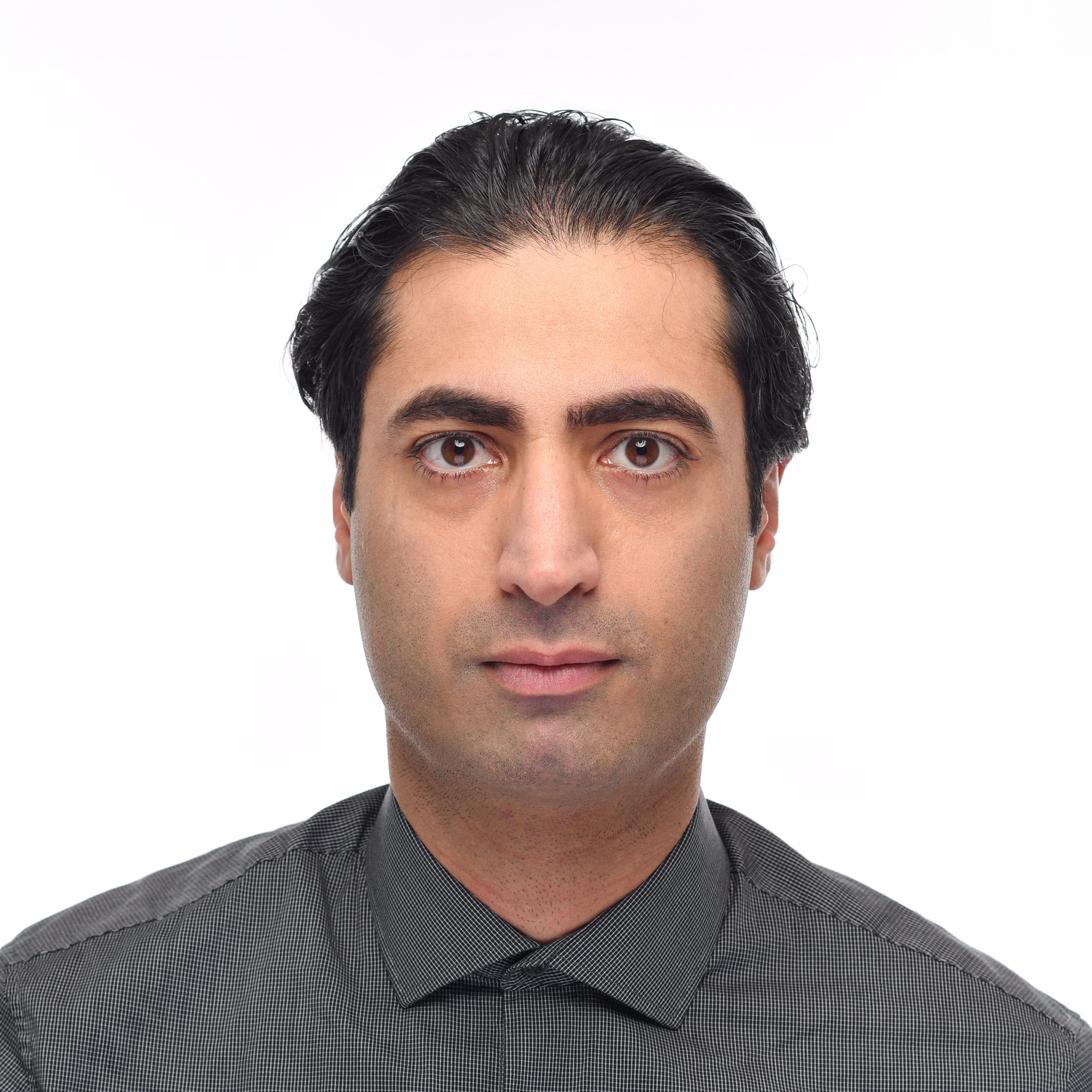}}]{Behnood Rasti}
(M’12–SM’19) received the B.Sc. and M.Sc. degrees in electronics and electrical engineering from the Electrical Engineering Department, University of Guilan, Rasht, Iran, in 2006 and 2009, respectively. He was a valedictorian of his M.Sc. class in 2009. He received his Ph.D. degree in Electrical and Computer Engineering from the University of Iceland, Reykjavik, Iceland, in 2014. From 2015 to 2016, he served as a postdoctoral researcher in the Electrical and Computer Engineering Department at the University of Iceland. He subsequently became a lecturer in the Center for Engineering Technology and Applied Sciences, Department of Electrical and Computer Engineering, from 2016 to 2019. Dr. Rasti was a Humboldt Research Fellow in 2020 and 2021, and a Principal Research Associate with Helmholtz Zentrum Dresden-Rossendorf (HZDR), Dresden, Germany, from 2022 to 2023. He is currently a Senior Research Scientist at the Faculty of Electrical Engineering and Computer Science, Technische Universität Berlin, and the Berlin Institute for the Foundations of Learning and Data, Berlin, Germany. His research interests include machine learning, deep learning, signal and image processing, remote sensing, Earth observation, and artificial intelligence. Dr. Rasti was a recipient of the Doctoral Grant of the University of Iceland Research Fund “The Eimskip University Fund” in 2013 and the “Alexander von Humboldt Research Fellowship Grant” in 2019. He serves as an Associate Editor for the IEEE Geoscience and Remote Sensing Letters (GRSL).
\end{IEEEbiography}

\begin{IEEEbiography}[{\includegraphics[width=1in,height=1.25in,clip,keepaspectratio]{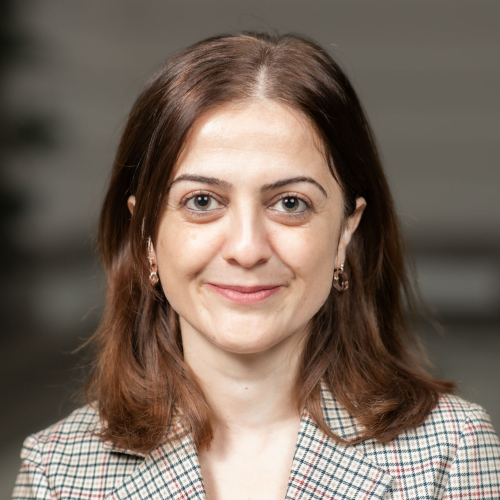}}]{Begüm Demir}
(S'06-M'11-SM'16) received the B.Sc., M.Sc., and Ph.D. degrees in electronic and telecommunication engineering from Kocaeli University, Kocaeli, Turkey, in 2005, 2007, and 2010, respectively. She  is currently a Full Professor and the founder head of the Remote Sensing Image Analysis (RSiM) group at the Faculty of Electrical Engineering and Computer Science, TU Berlin and the head of the Big Data  Analytics for Earth Observation research group at the Berlin Institute for the Foundations of Learning and Data (BIFOLD). Her research activities lie at the intersection of machine learning, remote sensing and signal processing. Specifically, she performs research in the field of processing and analysis of large-scale Earth observation data acquired by airborne and satellite-borne systems. She was awarded by the  prestigious ‘2018 Early Career Award’ by the IEEE Geoscience and Remote  Sensing Society for her research contributions in machine learning for information retrieval in remote sensing. In 2018, she received a Starting Grant from the European Research Council (ERC) for her project “BigEarth: Accurate and Scalable Processing of Big Data in Earth Observation”. She is an IEEE Senior Member and Fellow of European Lab for Learning and Intelligent Systems (ELLIS). Dr. Demir is a Scientific Committee member of several international conferences and workshops. She is a referee for several journals such as the PROCEEDINGS  OF THE IEEE, the IEEE TRANSACTIONS ON GEOSCIENCE AND REMOTE SENSING, the IEEE GEOSCIENCE AND REMOTE SENSING LETTERS, the IEEE TRANSACTIONS ON  IMAGE PROCESSING, Pattern Recognition, the IEEE TRANSACTIONS ON CIRCUITS AND SYSTEMS FOR VIDEO TECHNOLOGY, the IEEE JOURNAL OF SELECTED TOPICS IN SIGNAL PROCESSING, the International Journal of Remote Sensing), and several international conferences. Currently she is an Associate Editor for the IEEE GEOSCIENCE AND REMOTE SENSING MAGAZINE.
\end{IEEEbiography}

\end{document}